\documentclass[lettersize,journal]{IEEEtran}
\RequirePackage{xspace}
\usepackage{amsmath,amsfonts}
\usepackage{algorithmic}
\usepackage{algorithm}
\usepackage{array}
\usepackage[caption=false,font=normalsize,labelfont=sf,textfont=sf]{subfig}
\usepackage{textcomp}
\usepackage{stfloats}
\usepackage{amssymb}
\usepackage{booktabs}
\usepackage{url}
\usepackage{bbding}
\usepackage{bm}
\usepackage{verbatim}
\usepackage{graphicx}
 \usepackage{multirow} 
\usepackage{cite}
\usepackage[dvipsnames]{xcolor}
\usepackage{colortbl}
\definecolor{mypink2}{rgb}{.99,.96,.98}
\definecolor{mypink1}{rgb}{.99,.93,.98}
\definecolor{mypink}{rgb}{.99,.90,.98}
\definecolor{mygray}{rgb}{.95,.95,.95}

\usepackage{xspace} 
\usepackage{amsmath} 

\makeatletter
\DeclareRobustCommand\onedot{\futurelet\@let@token\@onedot}
\def\@onedot{\ifx\@let@token.\else.\null\fi\xspace}
\def\eg{\emph{e.g}\onedot} 
\def\ie{\emph{i.e}\onedot} 
 
 \def\vs{\emph{vs}\onedot}

\def\etal{\emph{et al}\onedot}
\makeatother

\begin{document}

\title{Learning Prompt with Distribution-Based Feature Replay for Few-Shot Class-Incremental Learning}

\author{Zitong Huang$^1$, Ze Chen$^2$ , Zhixing Chen$^1$, Erjin Zhou$^2$, Xinxing Xu$^3$\\
Rick Siow Mong Goh$^3$, Yong Liu$^3$,  Wangmeng Zuo$^1$\textsuperscript{\Envelope}, Chun-Mei Feng$^3$\textsuperscript{\Envelope}

\thanks{Zitong Huang, Zhixing Chen and Wangmeng Zuo are with the School of Computer Science and Technology, Harbin Institute of Technology, Harbin 150001, China (e-mail: cswmzuo@gmail.com). 

Ze Chen and Erjin Zhou are with the Megvii Research, Megvii Technology Limited, Beijing, China. 

Xinxing Xu, Rick Siow Mong Goh, Yong Liu and Chun-Mei Feng are with the Institute of High Performance Computing (IHPC), Agency for Science, Technology and Research (A*STAR), Singapore (e-mail: fengcm.ai@gmail.com)}
}

\markboth{Journal of \LaTeX\ Class Files,~Vol.~14, No.~8, August~2021}%
{Shell \MakeLowercase{\textit{et al.}}: A Sample Article Using IEEEtran.cls for IEEE Journals}


\maketitle

\begin{abstract}
%

Few-shot Class-Incremental Learning (FSCIL) aims to learn new classes with few examples while retaining knowledge of previously encountered ones.
Existing studies relied on pure visual networks, while in this paper we solved FSCIL by leveraging the pretrained vision-language model and propose a simple yet effective framework, named \textbf{L}earning \textbf{P}rompt with \textbf{Di}stribution-based \textbf{F}eature Replay (LP-DiF).
%
%
We observe that using CLIP for zero-shot evaluation significantly outperforms leading methods.
%
%
Then, prompt tuning is involved to further improve its adaptation ability, enabling continuous learning of session-specific knowledge.
%
%
To prevent the learnable prompt from forgetting old knowledge, we propose a pseudo-feature replay approach.
Specifically, we preserve old knowledge of each class by maintaining a feature-level Gaussian distribution with a diagonal covariance matrix, which is estimated by the features of training images and synthesized features generated from a VAE.
When progressing to a new session, pseudo-features are sampled from old-class distributions combined with training images of the current session to optimize the prompt, thus enabling the model to learn new knowledge while retaining old knowledge.
Experiments on prevalent benchmarks, \ie, CIFAR100, mini-ImageNet, CUB-200, and more challenging benchmarks, \ie SUN-397 and CUB-200$^*$ proposed in this paper showcase the superiority of LP-DiF, achieving new state-of-the-art (SOTA) in FSCIL. Code is publicly available at \url{https://github.com/1170300714/LP-DiF}.
\end{abstract}


\section{Introduction}
\label{sec:intro}
\IEEEPARstart{C}{lass-Incremental Learning}
(CIL)~\cite{wang2023comprehensive,zhou2023deep,de2021continual} faces challenges in data-scarce real-world applications, \eg, face recognition systems~\cite{zhao2023few} and smart photo albums~\cite{tao2020few}. 
This has led to the emergence of Few-Shot CIL (FSCIL)~\cite{tian2023survey}, where models adapt to new classes with limited training data, showcasing their relevance and flexibility in data-scarce scenarios.

In FSCIL, with only a few samples for each incremental task, the main challenge is not just avoiding catastrophic forgetting of previous knowledge~\cite{tao2020few,tian2023survey,song2023learning} but also facilitating plasticity from limited data.
Existing studies usually address this by first pre-training a classifier on a base set with numerous images for a robust foundation~\cite{shi2021overcoming,zhang2021few,zhu2021self,zhou2022few,ji2023memorizing,zhou2022forward,tao2020few,zhao2023few}.
Subsequent adaptations, \eg, knowledge distillation~\cite{zhao2023few}, class relationship modeling~\cite{tao2020few,zhang2021few}, and specific optimization~\cite{shi2021overcoming}, are then applied to the sparse incremental session data to boost performance while maintaining previously acquired knowledge.

\begin{figure}[t]

    \begin{center}
    \includegraphics[width=1\columnwidth]{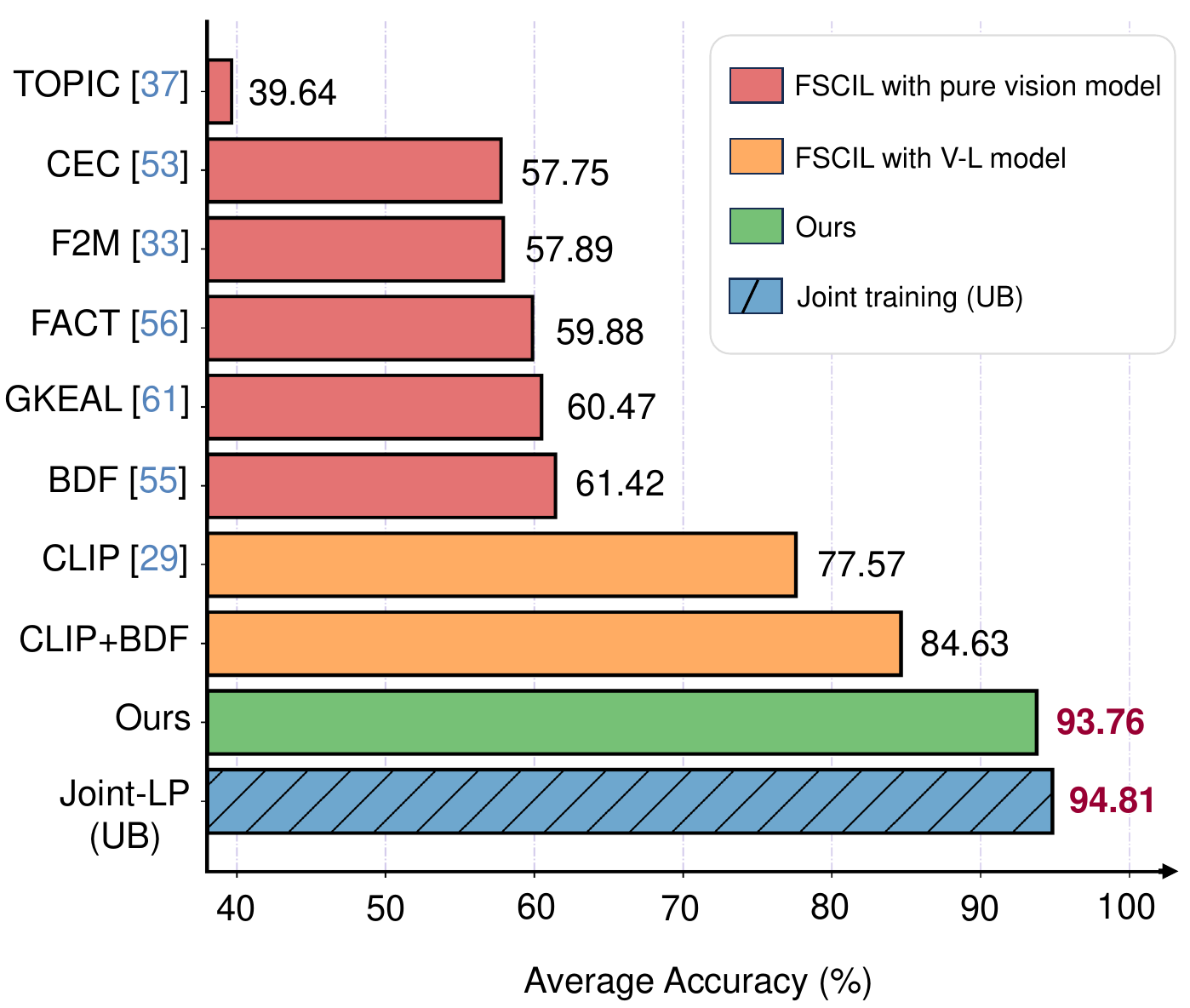}
    \end{center}

    \caption{\textbf{Comparison} of FSCIL methods in terms of Average Accuracy (\%) on the \texttt{test set} of \textbf{\textit{mini}-ImageNet} benchmark~\cite{russakovsky2015imagenet} under 5-shot setting. {{\texttt{Red}}}-highlighted bars indicate SOTA vision-based models (\eg, CNN~\cite{he2016deep}), while {{\texttt{orange}}} highlights show V-L pretrained models enhancing FSCIL, significantly outperforming those vision-based counterparts. Our method, marked in {{\texttt{green}}}, achieves 93.76\%, surpassing CLIP+BDF by 9.13\%, and comparable to the theoretical upper bound (UB) that highlights in {{\texttt{blue}}} achieved through learning prompt in joint-training manner.}
    \label{fig:intro_fig}

\end{figure}

This work diverges from approaches that solely rely on visual networks~\cite{he2016deep}, opting instead to leverage the capabilities of a Vision-Language (V-L) pretrained model, \ie, CLIP~\cite{radford2021learning,zhou2022learning}, to develop a few-shot incremental learner. 
Comparing with existing state-of-the-art techniques (see the {{\texttt{Red}}}-highlighted bars in Fig.~\ref{fig:intro_fig}), we observed that by simply crafting the manual prompt ``\texttt{A photo of a [CLS]}" as textual input and performing zero-shot evaluation on the widely used FSCIL benchmark, \textit{mini}-ImageNet~\cite{russakovsky2015imagenet} test set, CLIP (refer to the {\texttt{orange}} CLIP bar in Fig.~\ref{fig:intro_fig}) substantially outperforms all these SOTA methods, with a notable $16.15$\% performance boost over BiDistFSCIL (BDF)~\cite{zhao2023few}.
This finding indicates that the generalization abilities of V-L pretrained models are highly beneficial for FSCIL, \eg, naturally mitigating the plasticity issues caused by limited training samples.
Further, from Fig.~\ref{fig:intro_fig}, simply replacing the existing backbones of current SOTA methods with a pretrained image encoder, initializing and learning the classifier with the corresponding text encoding of manual prompt can further enhance performance (7.16\% gain to CLIP) but still lag behind the UB (9.13\% lower than the UB)  .
Therefore, how to derive an efficient and lightweight prompt for FSCIL continues to be a compelling challenge.

Based on the above preliminary results, this paper proposes a simple yet effective FSCIL framework by learning a lightweight prompt built upon the V-L pre-trained models.
Unlike CLIP, as well as simply integrating CLIP with existing methods (refer to the orange bar in Fig.~\ref{fig:intro_fig}, we resort to improving prompt tuning~\cite{zhou2022learning} for meeting the requirements of FSCIL.  
%
%
Specifically, for session $t$, we take the prompt in session $t-1$ for initialization, combine it with \texttt{[CLS]} to create the full-text input for each class, and then optimize learnable prompt with training data. 
%

To prevent the learnable prompt from forgetting prior knowledge in a new session, we also propose a pseudo-feature replay technique.
Specifically, observing that the image features extracted by the image encoder of CILP for each class seem to follow a Gaussian distribution (refer to Fig.~\ref{fig:gaosi}), we attempt to estimate its mean vector and diagonal covariance matrix (\ie parameters of Gaussian distribution) to fit the training data of each class.
%
To this end, a VAE \cite{kingma2013auto,wang2023improving} comprised of the V-L model and lightweight MLPs are proposed to synthesize features based on the few training samples and text information, permitting the usage of real image features as well as synthesized features to estimate Gaussian distribution parameters more accurately.
When the model trains on a new session, pseudo-image features from the old-class distributions are sampled as old-knowledge replay to constrain the optimization direction of the prompt, avoiding learning towards catastrophic forgetting.
The results in Fig.~\ref{fig:intro_fig} showcase that our approach improves zero-shot evaluation for CLIP by $16.19$\% and for CLIP+BDF by $9.13$\%. Notably, our method is merely $1.05$\% lower than the upper bound (Joint-LP, \ie, learning prompt on training data of each session jointly).

 In a nutshell, the main contributions of this paper are summarized as follows:

 \begin{itemize}

    \item [1)] We empirically show that pretrained V-L models, \eg CLIP, are beneficial for FSCIL due to its considerable generalization ability, inspiring us to propose a simple yet effective V-L based FSCIL method named LP-DiF.
   \item [2)] We adopt prompt tuning for allowing the model to continually capture specific knowledge of each session, and present a feature replay technique to prevent  catastrophic forgetting.
   %
   %
   By constructing feature-level Gaussian distribution for each class, pseudo feature replay can be combined with training images of current session to learn new knowledge while retaining old knowledge.
   \item [3)] Extensive evaluations and comparisons on three prevalent FSCIL benchmarks (CIFAR-100, CUB-200 and \textit{mini}-ImageNet) and two proposed more challenging benchmarks (SUN-397 and CUB-200$^*$) show the superiority of our methods in comparison to state-of-the-arts.
   %

\end{itemize}

\section{Related Work}
\noindent\textbf{Few-Shot Class-Incremental Learning.}
The few-shot class-incremental learning methods (FSCIL) aims to train a model in a class-incremental manner~\cite{de2021continual,zhou2023deep} with only a few samples for each new tasks~\cite{tian2023survey}.
Existing studies can be categorized into four families, \ie, dynamic network-based methods, meta-learning-based methods, feature space-based methods, and replay-based methods.
%
%
In specific, dynamic network structure~\cite{tao2020few,yang2021learnable,yang2022dynamic,gu2023few} is proposed to adaptive learn the new knowledge by dynamically expanding the network structure, so that the new knowledge is preserved by the new network structure.
Meta learning-based methods~\cite{mazumder2021few,zhu2021self,hersche2022constrained,chi2022metafscil,yang2023neural,zou2022margin,zhang2023controlvideo} employ a session sampling scheme, where a sequence of sessions are sampled from the base session, aiming to mimic the incremental learning process during evaluation, to allows the model to learn how to retain old knowledge under the condition of a small number of new data samples.
Feature space-based methods~\cite{cheraghian2021synthesized,akyurek2021subspace,kim2022warping,zhao2021mgsvf,zhou2022forward,zhou2022few,zhuang2023gkeal,ahmad2022few,ahmad2022variable}, focus on mapping the original image into a condensed feature space while preserving its essential attributes, which  ensures that the representations of old category data are not disrupted when the model is trained on new data. 
Replay-based methods~\cite{kukleva2021generalized,cheraghian2021semantic,dong2021few} retain or produce significant data from prior tasks to be reintroduced in the ongoing task. 
These methods are dedicated to selecting the most representative samples of old categories, or utilizing generative models to produce high-quality pseudo-samples of old categories.
While these methods have shown commendable performance, all those studies are based on feature extractors and classifiers built from deep networks trained in the base session.
Due to the scarcity of incremental class samples, the feature representation ability is limited.
%
%
In contrast, we propose to construct an incremental learner on a VL pre-trained model~\cite{radford2021learning,zhou2022learning} that offers inherent merits for FSCIL, \ie, endowing the image encoder with powerful feature representation abilities.

\vspace{5pt}
\noindent\textbf{Replay-based Incremental Learning.}
%
The replay-based approach in incremental learning leverages knowledge from previous tasks to mitigate catastrophic forgetting in models~\cite{chaudhry2018riemannian,bang2021rainbow,aljundi2019gradient,isele2018selective,rolnick2019experience,chaudhry2021using,shin2017continual,he2018exemplar,hu2018overcoming,agarwal2022semantics,chen2024saving}.
A basic data replay approach involves retaining a concise exemplar set, capturing essential samples from prior tasks~\cite{chaudhry2018riemannian,bang2021rainbow}, then, the classification model is trained on the combination of exemplars and the data of the current task.  
%
%
Different from directly storing the real instances, several following works~\cite{shin2017continual,he2018exemplar,hu2018overcoming,jiang2021ib} leveraged a generative model~\cite{goodfellow2014generative,kingma2013auto} for generating data from previous tasks.
%
Compared to methods based on real image replay, pseudo replay reduces storage needs by eliminating the requirement for exemplars and enriches the diversity of samples from previous tasks.
Yet, the overhead of training the image generator and dynamically producing pseudo images introduces additional computational demands and prolongs training time.
Instead of retaining an image generator, we represent the feature representation for each class using a Gaussian distribution, utilizing it to sample pseudo-features for rehearsing prior knowledge.
Moreover, drawing samples from this distribution is computationally efficient, offering our method an effective way for handling prevent catastrophic forgetting.
%

\begin{figure*}[t]
 
    \begin{center}
    \includegraphics[width=1.0\linewidth]{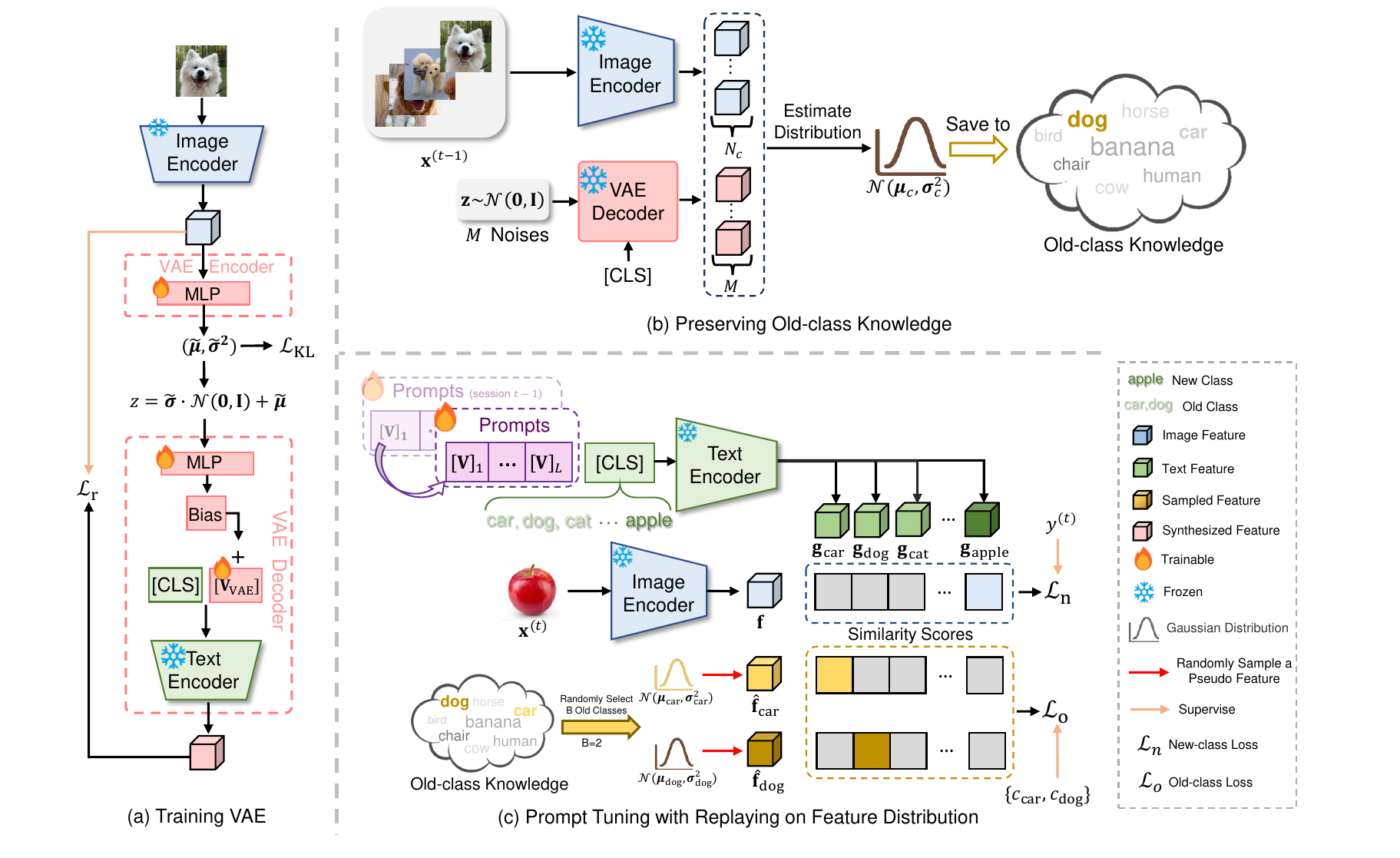}

    \end{center}

    \caption{\textbf{Overview }of our proposed \textbf{LP-DiF}. \textbf{(a)} In each session, we first train a VAE~\cite{kingma2013auto,wang2023improving} comprised of the V-L model and lightweight components, \ie, MLPs and learnable prompt, based on few training data and textual information of this session. \textbf{(b)} We preserve the knowledge of each class by estimating their feature-level statistical distribution. The mean vector and diagonal covariance matrix of the distribution are estimated by both the features of real images and the synthesized features from trained VAE. \textbf{(c)} Prompt is trained jointly with the combination of the real image of the current session and the pseudo-features sampled from old-class distributions.} 

    \label{fig:pipeline}
    \end{figure*}
    
\vspace{5pt}
\noindent\textbf{Incremental Learning via Pre-trained Model.}
%
Recent studies have explored constructing incremental learners using pre-trained models~\cite{douillard2022dytox,thengane2022clip,wang2022learning,wang2022dualprompt,smith2023coda,wang2023attriclip,wang2022s,yang2023continual,zhang2023slca}. The core idea of these methods is to leverage a pre-trained backbone, \eg, ViT~\cite{dosovitskiy2020image}, for robust image feature extraction, while only fine-tuning a selected set of parameters to adapt to new tasks.
For example, L2P~\cite{wang2022learning} employs a fixed pre-trained ViT as its backbone and sustains a dynamic prompt pool with various sets of adaptable prompts.
Some following works~\cite{wang2023attriclip,smith2023coda} built upon this concept, applying it to the VL pretrained model~\cite{radford2021learning}, leveraging linguistic knowledge to bolster classification performance.
In addition, Yang. \etal~\cite{yang2023continual} built a  Bayesian model based on a fixed feature extracted by a pretrained backbone. During the test stage, they use this Bayesian model as the classifier to mitigate the forgetting problem.
The above studies underscore the significant advantages of using pretrained models to boost performance in standard CIL scenarios.
As for FSCIL, we inherit the advantages of pretrained models in CIL.
Inspired by Yang. \etal~\cite{yang2023continual}, we further explore the potential characteristics of fixed feature, maintain a feature-level Gaussian distribution for each class to preserve the old knowledge, and use it to generate pseudo features to mitigate the catastrophic forgetting.

\section{Proposed Method}
\noindent\textbf{Problem Formulation.}
The purpose of FSCIL is to continually learn knowledge of new classes from few samples, while simultaneously preventing the model from forgetting knowledge of old classes.
Formally, a model is trained by a sequence of training data $\mathcal{D}_{\text{Train}} =\{D^{(t)}_{\text{Train}}\}^{T}_{t=0}$ continually, where $D^{(t)}_{\text{Train}}=\{(\mathbf{x}_i,y_i)\}^{N^{(t)}}_{i=0}$ denotes the training set of session (task) $t$. 
$\mathbf{x}_i$ is a training image with corresponding class label $y_i \in \mathcal{C}^{(t)}$, where $\mathcal{C}^{(t)}$ denotes the class space of $D^{(t)}_{\text{Train}}$.
For different sessions, the class spaces are non-overlapping, \ie $\forall t_1,t_2 \in \{0,1,\dots,T\}$ and $t_1 \neq t_2$, $\mathcal{C}^{(t_1)} \cap \mathcal{C}^{(t_2)} = \varnothing$.
Typically, $D^{(0)}_{\text{Train}}$ of the first session (\ie $t=0$), which is usually referred to as the base session, contains a substantial amount of training data.
While $D^{(t)}_{\text{Train}}(t>0)$ of the incremental sessions only contains few training sample, organized as the \textit{N}-Way \textit{K}-shot format, \ie, \textit{N} classes in each incremental session with each class comprising \textit{K} training images.
Following the formulation of standard class-incremental learning, in session $t$, only $D^{(t)}_{\text{Train}}$ and an optional memory buffer used to store the old knowledge (\eg exemplar) can be accessed.
After finishing training on $D^{(t)}_{\text{Train}}$, the model is evaluated on a test set $D^{(t)}_{\text{Test}}$, the class space of which is union of all the classes encountered so far, \ie $\mathcal{C}^{(0)} \cup \mathcal{C}^{(1)} \cdots 
 \cup \mathcal{C}^{(t)}$.

In this section, we propose a FSCIL method based on the V-L pretrained model, \eg CLIP~\cite{radford2021learning}.
We assume that the class names are accessible during the training and testing of each session. 
%
Formally, CLIP contains an image encoder $E_{\text{Img}}(\mathbf{x})$ and a text encoder $E_{\text{Txt}}(\mathbf{p})$, which are pretrained jointly with a huge amount of image-text pairs in contrastive learning manner.
An image $\mathbf{x}$ is fed into the image encoder, obtaining the corresponding $L_2$-normalized feature $\mathbf{f}$.
$\mathbf{p}$ is a text token which is obtained by tokenizing a sentence like ``\texttt{A photo of a [CLS]}.'', where [CLS] represents a certain class name.
We replace [CLS] by each class name respectively and obtain a set of text tokens $\{\mathbf{p}_c\}^{C}_{c=1}$, where $C$ denotes the total number of classes encountered so far. 
Then, $\{\mathbf{p}_c\}^{C}_{c=1}$ are fed into the text encoder, obtaining the corresponding $L_2$-normalized text feature $\{\mathbf{g}_c\}^{C}_{c=1}$.
Finally, the prediction score of class $c$ is computed by:

\begin{equation}
    p(y=c|\mathbf{x}) = \frac{\text{exp}(\langle \mathbf{f},\mathbf{g}_c\rangle/\tau)}{\sum^{C}_{j=1} \text{exp}(\langle\mathbf{f},\mathbf{g}_j\rangle/\tau)},
\end{equation}
where $\langle\cdot,\cdot\rangle$ denotes the cosine similarity of the two features and $\tau$ is the temperature parameter.

\begin{figure*}[t]
    \begin{center}
   
    \includegraphics[width=1.0\linewidth]{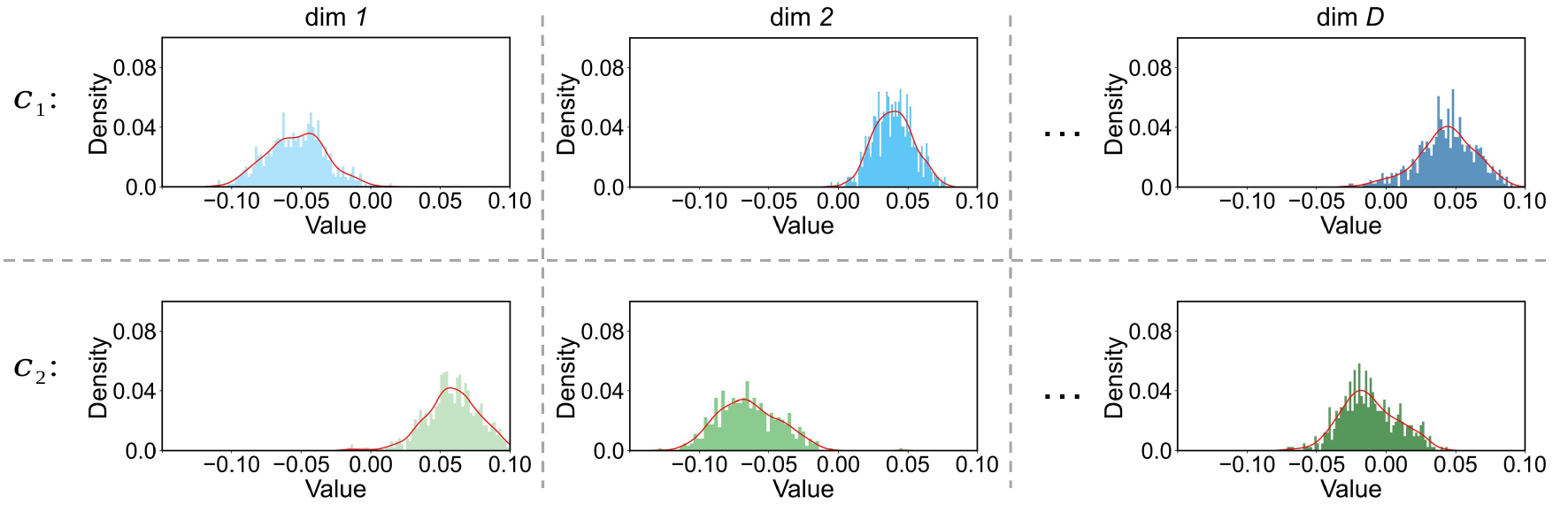}
    \end{center}

    \caption{\textbf{Histogram visualization} of the \textit{statistical distribution} of image features. We take the image features with different dimensions (dim) of classes $c_1$ and $c_2$ as example selected from the {\textit{mini}-ImageNet}~\cite{russakovsky2015imagenet} benchmark by the image encoder of CLIP (ViT-B/16)~\cite{radford2021learning}. 
    Each sub-figure shows the distribution with histogram of corresponding random variable $Z_{cd}$, where $c$ and $d$ denotes the index of class and feature dimension respectively.
    Obviously, \textbf{1)} each dimension of the image features per class approximates Gaussian distribution; \textbf{2)} distributions of same dimension vary in different classes, \eg, $Z_{c_{1}1}$ \textit{{vs.}} $Z_{c_{2}1}$ .} 
    \label{fig:gaosi}
\end{figure*}

\subsection{Approach Overview}\label{sec:overview}
Although CLIP has demonstrated its superior performance on FSCIL in Fig.~\ref{fig:intro_fig}, using hand-crafted prompt is sub-optimal for transfer the knowledge to each incremental session.
%
%
So we replace the hand-crafted prompt with a set of learnable vectors $\mathcal{V} =\{[\mathbf{V}]_l\}^L_{l=1}$~\cite{zhou2022learning}, 
where $[\mathbf{V}]_l$ ($l \in \{1,\dots,L\}$) denotes one learnable vector, and $L$ is the number of vectors.
Hence, the expression for the text prompt is modified to:
\begin{equation}
    \mathbf{p}(\mathcal{V}) = [\mathbf{V}]_1 [\mathbf{V}]_2 \dots [\mathbf{V}]_L [\text{CLS}],
\end{equation}
To learn $\mathcal{V}$ on $\mathcal{D}_\text{Train}$, an intuitive approach is to sequentially tune the prompt using training data from each incremental session to continually acquire new knowledge.
Specifically, at the beginning of session $0$, we initialize $\mathcal{V}$  randomly;
while for each following session $t$ $(t>0)$, we use the $\mathcal{V}$ trained in the previous session (\eg session $t-1$) to initialize the $\mathcal{V}$ for current session.
In a certain session $t$, given a pair of training sample $(\mathbf{x}_i,y_i)$ from $D^{(t)}_{\text{Train}}$, prompt is optimized on by minimizing $\mathcal{L}_\text{n}$:
\begin{equation}\label{eq:l_n}
    \mathcal{L}_\text{n} = -\log \frac{\text{exp}(\langle\mathbf{f}_i, E_{\text{Txt}}(\mathbf{p}_{y_i}(\mathcal{V}))\rangle/\tau)}{\sum^{\left|\bigcup^{t}_{s=0} \mathcal{C}^{(s)}\right|}_{c=1} \text{exp}(\langle\mathbf{f}_i,E_{\text{Txt}}(\mathbf{p}_{c}(\mathcal{V}))\rangle/\tau)},
\end{equation}
where $\mathbf{f}_i$ denotes the $L_2$-normalized image feature of $\mathbf{x}_i$, and $\mathbf{p}_c(\mathcal{V})$ denotes the prompt corresponding to class $c$.
%


However, using only the $\mathcal{D}^{(t)}_\text{Train}$  to optimize the prompt in session $t$ will inevitably lead to catastrophic forgetting.
%
Ideally, learning prompt with all training data from both previous and current sessions (\eg $\bigcup^{t}_{s=0}\mathcal{D}^{(s)}_\text{Train}$) can address this issue, but this is not allowed under the protocol of FSCIL.
Therefore,  this paper adopts a compromise solution, proposing to record old knowledge by maintaining statistical distributions of old classes instead of directly storing origin images.
%
%
 We setup a feature-level Gaussian distribution to represent each old class, which is represented by a mean vector and a diagonal covariance matrix.
We name it the \textbf{old-class distribution}.
The mean vector and diagonal covariance matrix of the old-class distribution are estimated jointly from the features of real images as well as synthetic features generated by a VAE decoder.
%
When learning the prompt in a new session, we randomly sample features based on the statistical distribution of old classes to replay old knowledge. 
Then, the sampled features of old classes and the real features of new classes will jointly optimize the prompt, thereby learning new knowledge while also replaying old knowledge.
%
In the following, We will introduce how to obtain the old-class distribution in Sec~\ref{sec:construct}, and how to learn prompt in Sec~\ref{sec:replay}.

\subsection{Estimation of Old-Class Distribution}\label{sec:construct}
In each session $t$, we should estimate the feature-level statistical distribution for each class of $\mathcal{D}^{(t)}_\text{Train}$. 
%
Given a certain class label $c \in \mathcal{C}^{(t)}$ , the corresponding training images $\{\mathbf{x}_i\}^{N_c}_{i=1}$ are fed into the image encoder $E_\text{Img}(\mathbf{x})$ to obtain their $L_2$-normalized features $\{\mathbf{f}_i\}^{N_c}_{i=1}$, where $N_c$ denotes the number of training images of class $c$, $\mathbf{f}_i=[f_{i1},f_{i2},\dots,f_{iD}]^T$ and $D$ is the feature dimension (\eg $D=512$ for ViT-B/16).
Intuitively, we assume that the features of class $c$ follow a multivariate distribution $\mathcal{N}(\bm{\mu}_c, \bm {\Sigma}_c)$, where $\bm{\mu}_c \in \mathbb{R}^D$ denotes the mean vector and $\bm {\Sigma}_c \in \mathbb{R}_{\geq 0}^{D \times D}$ denotes the covariance matrix.
%
As shown in Fig.~\ref{fig:gaosi}, we observe that each dimension of these features of each class approximates a Gaussian distribution, and distributions of same dimension
vary in different classes. 
%
%
Thus, each dimension of the feature can be treated as independently distributed, and the covariance matrix $\bm {\Sigma}_c$ can be simplified to a diagonal matrix and be represented by a vector $\bm{\sigma}_{c}^2=[\sigma_{c1}^2,\sigma_{c2}^2,\dots,\sigma_{cD}^2]^T$, which is diagonal values of $\bm {\Sigma}_c$.
%
%
We use random variable $Z_{cd}$ to represent the $d$-th dimension of feature, following a specific Gaussian distribution $\mathcal{N}(\mu_{cd},\sigma^2_{cd})$, where $\mu_{cd}$ denotes the mean value of the $d$-th dimension.
Then, $\mathbf{Z}_{c} = [Z_{c1},Z_{c2},\dots,Z_{cD}]$ represents the random variable of the whole feature following $\mathcal{N}(\bm{\mu}_c,\bm{\sigma}_{c}^2)$.
%
Our goal is to estimate the $\bm{\mu}_c$ and $\bm{\sigma}_{c}^2$ for each class.

For each class, simply using only $\{\mathbf{f}_i\}^{N_c}_{i=1}$  to estimate the $\bm{\mu}_c$ and $\bm{\sigma}_c^2$ may be inadequate due to the scarcity of the data. 
To tackle with this problem, we utilize a VAE~\cite{kingma2013auto,wang2023improving} comprised of the V-L models and lightweight MLPs, leveraging the few training data and textual information to synthesize more image features, thereby benefiting the estimation of the distribution.
As shown in Fig.~\ref{fig:pipeline} (a), in VAE Encoder, an image feature $\mathbf{f}$ is fed into a MLP, encoded to a latent code $\mathbf{z}$, of which distribution is assumed to be a prior $\mathcal{N}(\mathbf{0},\mathbf{I})$:
\begin{equation}
    \mathcal{L}_\text{KL} = KL(\mathcal{N}(\tilde{\bm{\mu}},\tilde{\bm{\sigma}}^2)||\mathcal{N}(\mathbf{0},\mathbf{I})),
\end{equation}
where $KL$ represents the Kullback-Leibler divergence.
In VAE Decoder,  $\mathbf{z}$ is fed to another MLP and obtain the bias $\bm r$, which is added to a set of learnable prompt $\mathcal{V}_\text{VAE}=\{[\mathbf{V}_\text{VAE}]_l\}^L_{l=1}$:
\begin{equation}
    \mathcal{V}_\text{VAE}(\mathbf{z}) = \{[\mathbf{V}_\text{VAE}]_l+\bm r\}^L_{l=1}.
\end{equation}
Then, $\mathcal{V}_\text{VAE}(\mathbf{z})$ concatenating with the class name $[\text{CLS}]$ corresponding to $\mathbf{f}$ is fed into the text encoder, obtaining the reconstruct feature $\mathbf{\tilde{f}}$ then calculating the reconstruct loss $\mathcal{L}_\text{r}$:
\begin{equation}
    \mathcal{L}_\text{r} = \lVert \mathbf{f}-\mathbf{\tilde{f}} \rVert_2.
\end{equation}
Finally, the total loss $\mathcal{L}_\text{VAE}$ of training the VAE is:
\begin{equation}
    \mathcal{L}_\text{VAE} = \mathcal{L}_\text{KL}+\lambda_\text{r}\mathcal{L}_\text{r},
\end{equation}
where $\lambda_\text{r}$ represents the coefficient of $\mathcal{L}_\text{r}$.

Using both the features synthesized by the VAE and the real image features, we estimate $\bm{\mu}_c$ and $\bm{\sigma}_c^2$.
As shown in Fig.~\ref{fig:pipeline} (b), for a specific class $c$, 
$M$ noise vectors $\mathbf{z} \sim \mathcal{N}(\mathbf{0},\mathbf{I})$ and corresponding class name are input into the VAE Decoder, obtaining $M$ synthesized features $\{\mathbf{\tilde{f}}_j\}^{M}_{j=1}$.
Then, $\bm{\mu}_c$ and $\bm{\sigma}_{c}^2=[\sigma_{c1}^2,\sigma_{c2}^2,\dots,\sigma_{cD}^2]^T$ are estimated by:
\begin{equation}
    \bm{\mu}_c = \frac{1}{N_c+M}(\sum^{N_c}_{i=1} \mathbf{f}_i +\sum^{M}_{j=1} \mathbf{\tilde{f}}_j),
\end{equation}
%
\begin{equation}
\small
\sigma_{cd}^2 = \frac{1}{(N_c+M)-1}(\sum_{i=1}^{N_c} (f_{id} - \mu_{cd})^2+\sum_{j=1}^{M} (\tilde{f}_{jd} - \mu_{cd})^2).
\end{equation}

%
\subsection{Learning Prompt with Feature Replay}\label{sec:replay}
At session $t$, we learn prompt with  $\mathcal{D}^{(t)}_\text{Train}$ as well as the distributions of old classes preserved in previous sessions. 
 %
 %
%
 \begin{itemize}
    \item [1)]  When $t=0$, \ie, the first session, we just follow the approach in Sec.\ref{sec:overview}, randomly initializing $\mathcal{V}$ and learning them with $\mathcal{D}^{(0)}_\text{Train}$ by $\mathcal{L}_\text{n}$ (Eq.~(\ref{eq:l_n})).
   \item [2)] When $t>0$, $\mathcal{V}$ are initialized from trained weights in session $t-1$. 
   %
   For the new knowledge of $\mathcal{D}^{(t)}_\text{Train}$, we adopt $\mathcal{L}_\text{n}$.
   For the old knowledge of previous sessions,
    we randomly sample pseudo image features of old classes from their corresponding distributions.
   %
  As shown in Fig.\ref{fig:pipeline} (c), for each selected training image in one batch $\mathbf{x_i}$, we first randomly select $B$ old classes: $\{c_b\}^{B}_{b=1}$ and $c_b \in \cup^{t-1}_{s}\mathcal{C}^{(s)}$.
   Then, for each selected class $c_b$, we randomly sample a pseudo feature $\mathbf{\hat{f}}_{c_b}$ from its feature distribution $ \mathbf{\hat{f}}_{c_b} \sim \mathcal{N}(\bm \mu_{c_b}, \bm{\sigma}^2_{c_b})$.
   %
   These sampled features with their corresponding class labels are used to calculate the loss $\mathcal{L}_\text{o}$:
\begin{equation}\label{eq:l_o}
    \mathcal{L}_\text{o} = -\sum^{B}_{b=1}\log \frac{\text{exp}(\langle\mathbf{\hat{f}}_{c_b}, E_{\text{Txt}}(\mathbf{p}_{c_b}(\mathcal{V}))\rangle/\tau)}{\sum^{\left|\bigcup^{t}_{s=0} \mathcal{C}^{(s)}\right|}_{c=1} \text{exp}(\langle \mathbf{\hat{f}}_{c_b},E_{\text{Txt}}(\mathbf{p}_{c}(\mathcal{V}))\rangle/\tau)}.
\end{equation}
\end{itemize}
Finally, the prompt is optimized by minimizing the loss:
\begin{equation}
    \mathcal{L}_\text{LP} =
\begin{cases}
\mathcal{L}_\text{n} & \text{if } t = 0,\\
\mathcal{L}_\text{n} + \lambda_\text{o} \mathcal{L}_\text{o} & \text{if } t > 0,
\end{cases}
\end{equation}
where $\lambda_\text{o}$ represents the tradeoff coefficient.

\section{Experiments}
\label{sec:exp}

\begin{table*}[!htb] 
	\begin{center}

		\caption{ \textbf{Comparison} on {\textit{mini}-ImageNet}. ``\textbf{Backbone.}'' represents the backbone of visual model. ``\textbf{Avg}'' represents the average accuracy of all sessions; the higher the value, the better performance. ``\textbf{PD}'' represents the Performance Drop rate; the lower the value, the better performance. ``\textbf{bs}'' is the abbreviation of ``baseline''.  ``\textbf{\textit{UB.}}'' is the abbreviation of ``Upper bound''.}
 \fontsize{8}{10}\selectfont
 \setlength{\tabcolsep}{7.5pt}
			\begin{tabular}{lccccccccccccc}
				\toprule
				\multicolumn{1}{l}{\multirow{2}{*}{\textbf{Methods}}} & 
    \multicolumn{1}{c}{\multirow{2}{*}{\textbf{Backbone}}} &
    \multicolumn{9}{c}{\textbf{Accuracy in each session (\%)} $\uparrow$} & \multicolumn{1}{l}{\multirow{2}{*}{\textbf{Avg} $\uparrow$}} & \multicolumn{1}{l}{\multirow{2}{*}{\textbf{PD} $\downarrow $}}  \\ 
				\cmidrule{3-11}
				& &  0   &  1      &  2      &  3    &  4     &  5  & \ 6     &  7      & 8   &  &  \\ 
\midrule
                
				TOPIC~\cite{tao2020few}  
                & Res18
                &$61.31$	
                &$50.09$	
                &$45.17$	
                &$41.16$	
                &$37.48$	
                &$35.52$	
                &$32.19$	
                &$29.46$	
                &$24.42$	
                &$39.64$ 
                &$36.89$\\

				CEC~\cite{zhang2021few}  
                & Res18
                &$72.00$	
                &$66.83$
                &$62.97$	
                &$59.43$	
                &$56.70$	
                &$53.73$	
                &$51.19$	
                &$49.24$	
                &$47.63$	
                &$57.75$
                &$24.37$\\
                
                F2M~\cite{shi2021overcoming} 
                & Res18
                &$72.05$ 
                &$67.47$ 
                &$63.16$ 
                &$59.70$ 
                &$56.71$ 
                &$53.77$ 
                &$51.11$ 
                &$49.21$ 
                &$47.84$ 
                &$57.89$ 
                &$24.21$ \\
				Replay~\cite{liu2022few} 
                & Res18
                &$71.84$	
                &$67.12$	
                &$63.21$	
                &$59.77$	
                &$57.01$	
                &$53.95$	
                &$51.55$	
                &$49.52$	
                &$48.21$	
                &$58.02$ 
                &$23.63$\\
                
				MF~\cite{chi2022metafscil} 
                & Res18
                &$72.04$
                &$67.94$
                &$63.77$
                &$60.29$
                &$57.58$
                &$55.16$
                &$52.90$
                &$50.79$
                &$49.19$
                &$58.85$
                &$22.85$ \\

                GKEAL~\cite{zhuang2023gkeal} 
                & Res18
                &$73.59$
                & $68.90$ 
                & $65.33$ 
                &$62.29$ 
                & $59.39$ 
                & $56.70$ 
                &$54.20$ 
                &$52.59$ 
                &$51.31$ 
                &$60.48$ 
                &$22.28$ \\
     
                FACT~\cite{zhou2022forward} 
                & Res18
                &$72.56$ 
                &$69.63$ 
                &$66.38$ 
                &$62.77$ 
                &$60.60$ 
                &$57.33$ 
                &$54.34$ 
                &$52.16$ 
                &$50.49$ 
                &$60.70$ 
                &$22.07$ \\

                C-FSCIL~\cite{hersche2022constrained}
                & Res12
                &$76.40$
                &$71.14$ 
                &$66.46$ 
                &$63.29$ 
                &$60.42$ 
                &$57.46$ 
                &$54.78$ 
                &$53.11$ 
                &$51.41$ 
                &$61.59$ 
                &$14.99$ \\
                
                BDF~\cite{zhao2023few} 
                & Res18
                &$74.65$ 
                &$70.70$ 
                &$66.81$ 
                &$63.63$ 
                &$61.36$ 
                &$58.14$ 
                &$55.59$ 
                &$54.23$ 
                &$53.39$ 
                &$62.06$ 
                &$21.26$ \\
                
                FCIL~\cite{gu2023few} 
                & Res18
                &$76.34$ 
                &$71.40$ 
                &$67.10$ 
                &$64.08$ 
                &$61.30$ 
                &$58.51$ 
                &$55.72$ 
                &$54.08$ 
                &$52.76$ 
                &$62.37$ 
                &$23.58$ \\
                
                SAVC~\cite{song2023learning} 
                & Res18
                &$81.12$ 
                &$76.14$ 
                &$72.43$ 
                &$68.92$ 
                &$66.48$ 
                &$62.95$ 
                &$59.92$ 
                &$58.39$ 
                &$57.11$ 
                &$67.05$ 
                &$24.01$\\
                
                NC-FSCIL~\cite{yang2022neural}  
                & Res18
                &$84.02$
                &$76.80$
                &$72.00$
                &$67.83$
                &$66.35$	
                &$64.04$
                &$61.46$
                &$59.54$
                &$58.31$
                &$67.82$
                &$25.71$ \\
   \cmidrule(lr){1-1}  \cmidrule(lr){2-10} \cmidrule(lr){11-12} 

                CLIP (\textit{Bs.})~\cite{radford2021learning} 
                & ViT-B/16
                &$80.01$ 
                &$79.16$ 
                &$78.89$ 
                &$77.97$ 
                &$77.44$ 
                &$76.83$ 
                &$76.32$ 
                &$76.02$ 
                &$75.45$ 
                &$77.57$
                &$\textbf{4.56}$ \\

                \textbf{LP-DiF (\textit{Ours})} 
                & ViT-B/16
                &$\textbf{96.34}$ 
                &$\textbf{96.14}$ 
                &$\textbf{94.62}$ 
                &$\textbf{94.37}$ 
                &$\textbf{94.06}$
                &$\textbf{93.44}$ 
                &$\textbf{92.21}$ 
                &$\textbf{92.29}$ 
                &$\textbf{91.68}$ 
                &$\textbf{93.76}$ 
                &$4.66$\\

\midrule
\midrule

                {Joint-LP (\textit{UB. of ours})} 
                & {ViT-B/16}
                &${{96.34}}$ 
                &${{96.07}}$ 
                &${{95.75}}$
                &${{94.93}}$
                &${{94.61}}$
                &${{94.26}}$
                &${{93.99}}$
                &${{93.83}}$
                &${{93.56}}$
                &${{94.81}}$
                &${{2.78}}$ \\

				\bottomrule
			\end{tabular}
		\label{table:imgnet}
	\end{center}
\end{table*}

\begin{table}[t] 

		\caption{\textbf{Details of selected benchmarks.} The first three lines are commonly used benchmarks, while the last two lines are the more challenging benchmarks proposed in this paper. $\bm{|\mathcal{C}^\text{All}|}$, $\bm{|\mathcal{C}^\text{Base}|}$ and $\bm{|\mathcal{C}^\text{Inc}|}$ denotes the total number of classes, the number of classes in base session, and the number of classes in each incremental session respectively. \textbf{\#Base} and \textbf{\#Inc} denote the number of base sessions and the incremental session respectively. \textbf{Shot} denotes the number of training images of each incremental session. $^*$ represents a variant version.}
        \fontsize{11}{16}\selectfont
		\resizebox{0.48\textwidth}{!}{
			\begin{tabular}{l|ccccccc}
				\toprule
				\textbf{Dataset} 
                &$\bm{|\mathcal{C}^\text{All}|}$
                &$\bm{|\mathcal{C}^\text{Base}|}$
                &$\bm{|\mathcal{C}^\text{Inc}|}$
                &\textbf{\#Base}
                &\textbf{\#Inc} 
                &\textbf{Shot}\\

\midrule
                CIFAR-100~\cite{krizhevsky2009learning}
                &100 
                &60
                &5
                &1
                &8 
                &5\\

                \textit{mini}-ImageNet~\cite{russakovsky2015imagenet} 
                &100 
                &60
                &5
                &1
                &8 
                &5\\

                CUB-200~\cite{wah2011caltech}
                &200 
                &100 
                &10
                &1
                &10
                &5\\

                \hline
                SUN-397~\cite{xiao2010sun}
                &397
                &197
                &10
                &1
                &20
                &5\\

                CUB-200*~\cite{wah2011caltech}
                &200
                &0
                &10
                &0
                &20
                &5\\

				\bottomrule
			\end{tabular}
		}
		\label{table:dataset_details}

\end{table}

\subsection{Datasets and Metrics}
\noindent\textbf{Datasets.} Following the mainstream benchmark settings~\cite{zhao2023few}, we conduct experiments on three datasets, \ie, {CIFAR-100}~\cite{krizhevsky2009learning}, {\textit{mini}-ImageNet}~\cite{russakovsky2015imagenet} and {CUB-200}~\cite{wah2011caltech}, to evaluate our LP-DiF. 
Tab.~\ref{table:dataset_details} summarizes the details of each selected benchmark.
\begin{itemize}
    \item CIFAR-100 dataset consists of $100$ classes, each of which contains $50,000$ training images.
Following the previous study~\cite{zhao2023few}, there are $60$ classes in the base session, and the remaining classes will be divided into $8$ incremental sessions, with each incremental session comprising $5$ classes.
    \item CUB-200 is a fine-grained classification dataset containing $200$ bird species with about $6,000$ training images.
Following the previous study~\cite{zhao2023few}, there are $100$ classes in the base session, and the remaining classes will be divided into $10$ incremental sessions, with each incremental session comprising $10$ classes.
    \item \textit{mini}-ImageNet is a smaller part of ImageNet~\cite{russakovsky2015imagenet}, which has $50,000$ training images from $100$ chosen classes.
Following the previous study~\cite{zhao2023few}, there are $60$ classes in the base session, and the remaining classes will be divided into $8$ incremental sessions, with each incremental session comprising $5$ classes.
\end{itemize} 

Additionally, this paper also proposes two more challenging benchmarks for FSCIL, \ie, {SUN-397}~\cite{xiao2010sun} and {CUB-200$^*$}.

\begin{itemize}
    \item SUN-397 is a large-scale scene understanding dataset containing $397$ distinct scene classes with about $76,000$ training images.
    We select $197$ classes for the base session; the remaining classes will be split into $20$ incremental sessions, with each incremental session comprising $10$ classes.
    %
\emph{We evaluate our method on this benchmark to reveal whether it is effective in scenarios with more classes and more incremental sessions.}
    \item {CUB-200$^*$} is a variant of CUB-200 but excludes the base session. 
We evenly divide the total $200$ classes into $20$ incremental sessions, with each session containing $10$ categories.
    Following the previous study~\cite{zhao2023few}, there are $100$ classes in the base session, and the remaining classes will be divided into $10$ incremental sessions, with each incremental session comprising $10$ classes. 
    \emph{We use it to evaluate whether our method works in scenarios without the base session.} 
\end{itemize} 
%
%
%
%

\noindent\textbf{Metrics.}
Following existing FSCIL methods~\cite{zhao2023few,song2023learning,tao2020few}, we employ the \textbf{Avg.}, which is the average accuracy of each session, as primary metric for performance comparison.
In addition, we also employ the \textbf{performance drop rate} (PD.), which represents the drop of performance of the last session compared to the first session, to reflect the extent of the model's forgetting of old knowledge.

\subsection{Implementation Details.}
All experiments are conducted with PyTorch on $8\times$ NVIDIA RTX $2080$Ti GPUs.
We leverage the ViT-B/16 as the image encoder of LP-DiF and adopt
SGD with $0.9$ momentum to optimize the prompts.
The learning rate is initialized by $0.002$.
For the base session, the batch size is set to $64$ and the training epoch is set to $200$,
As for each incremental session, the batch size and the training epochs are set to $25$, $100$, respectively.
The VAE component is enabled only for incremental sessions.
%
For the hyper-parameters, $M$ is set to $10$; $B$ and $\lambda_\text{o}$ are set to $8$ and $2$, respectively; 
$L$ is set to $16$ following Zhou~\etal~\cite{zhou2022learning};
$\lambda_\text{r}$ is set to $1$ following Wang~\etal~\cite{wang2023improving}.

\begin{table*}[!htb] 
	\begin{center}
        
		\caption{\textbf{Comparison} with state-of-the-art FSCIL methods on {CUB-200}. ``\textbf{Backbone.}'' represents the backbone of visual model. ``\textbf{Avg}'' represents the average accuracy of all sessions; the higher the value, the better performance. ``\textbf{PD}'' represents the Performance Drop rate; the lower the value, the better performance. ``\textbf{bs}'' is the abbreviation of ``baseline''.  ``\textbf{\textit{UB.}}'' is the abbreviation of ``Upper bound''.}

   \fontsize{8}{10}\selectfont
 \setlength{\tabcolsep}{5.3pt}
			\begin{tabular}{lccccccccccccccc}
				\toprule
				\multicolumn{1}{l}{\multirow{2}{*}{\textbf{Methods}}} & \multicolumn{1}{c}{\multirow{2}{*}{\textbf{Backbone}}} & \multicolumn{11}{c}{\textbf{Accuracy in each session (\%)} $\uparrow$} & \multicolumn{1}{l}{\multirow{2}{*}{\textbf{Avg} $\uparrow$}} & \multicolumn{1}{l}{\multirow{2}{*}{\textbf{PD} $\downarrow $}}  \\ 
				\cmidrule{3-13}
				&  & 0   &  1      &  2      &  3    &  4     &  5  & \ 6     &  7      & 8   & 9 & 10 & & \\ 
\midrule
                
				TOPIC~\cite{tao2020few}
                & Res18
                &$68.68$	
                &$62.49$	
                &$54.81$	
                &$49.99$	
                &$45.25$	
                &$41.40$	
                &$38.35$	
                &$35.36$	
                &$32.22$	
                &$28.31$ 
                &$26.26$
                &$43.92$ 
                &$42.42$\\
    
				CEC~\cite{zhang2021few}
                & Res18
                &$75.85$	
                &$71.94$
                &$68.50$	
                &$63.50$	
                &$62.43$	
                &$58.27$	
                &$57.73$	
                &$55.81$	
                &$54.83$	
                &$53.52$
                &$52.28$
                &$61.33$
                &$23.57$\\
                
				Replay~\cite{liu2022few}
                & Res18
                &$75.90$	
                &$72.14$	
                &$68.64$	
                &$63.76$	
                &$62.58$	
                &$59.11$	
                &$57.82$	
                &$55.89$	
                &$54.92$	
                &$53.58$ 
                &$52.39$
                &$61.52$ 
                &$23.51$\\
                
				MetaFSCIL~\cite{chi2022metafscil} 
                & Res18
                &$75.90$
                &$72.41$
                &$68.78$
                &$64.78$
                &$62.96$
                &$59.99$
                &$58.30$
                &$56.85$
                &$54.78$
                &$53.82$
                &$52.64$
                &$61.93$
                &$23.26$\\

                FACT~\cite{zhou2022forward}
                & Res18
                &$75.90$ 
                &$73.23$ 
                &$70.84$ 
                &$66.13$ 
                &$65.56$ 
                &$62.15$ 
                &$61.74$ 
                &$59.83$ 
                &$58.41$ 
                &$57.89$ 
                &$56.94$
                &$64.42$ 
                &$18.96$ \\

                FCIL~\cite{gu2023few}
                & Res18
                &$78.70$ 
                &$75.12$ 
                &$70.10$ 
                &$66.26$ 
                &$66.51$ 
                &$64.01$ 
                &$62.69$ 
                &$61.00$ 
                &$60.36$ 
                &$59.45$ 
                &$58.48$
                &$65.70$ 
                &$20.22$ \\

                GKEAL~\cite{zhuang2023gkeal}
                & Res18
                &$78.88$
                & $75.62$ 
                & $72.32$ 
                &$68.62$ 
                & $67.23$ 
                & $64.26$ 
                &$62.98$ 
                &$61.89$ 
                &$60.20$ 
                &$59.21$ 
                &$58.67$
                &$66.35$ 
                &$20.21$\\

                NC-FSCIL~\cite{yang2022neural}
                & Res18
                &$80.45$
                &$75.98$
                &$72.30$
                &$70.28$
                &$68.17$	
                &$65.16$
                &$64.43$
                &$63.25$
                &$60.66$
                &$60.01$
                &$59.44$
                &$67.28$
                &$21.01$ \\

                BiDistFSCIL~\cite{zhao2023few} 
                & Res18
                &$79.12$ 
                &$75.37$ 
                &$72.80$ 
                &$69.05$ 
                &$67.53$ 
                &$65.12$ 
                &$64.00$ 
                &$63.51$ 
                &$61.87$ 
                &$61.47$ 
                &$60.93$
                &$67.34$
                &$18.19$\\

                SAVC~\cite{song2023learning}
                & Res18
                &$81.85$ 
                &$77.92$ 
                &$74.95$ 
                &$70.21$ 
                &$69.96$ 
                &$67.02$ 
                &$66.16$ 
                &$65.30$ 
                &$63.84$ 
                &$63.15$ 
                &$62.50$
                &$69.35$ 
                &$19.35$ \\

                F2M~\cite{shi2021overcoming}
                & Res18
                &$81.07$ 
                &$78.16$ 
                &$75.57$ 
                &$72.89$ 
                &$70.86$ 
                &$68.17$ 
                &$67.01$ 
                &$65.26$ 
                &$63.36$ 
                &$61.76$ 
                &$60.26$
                &$69.49$ 
                &$20.81$ \\
   \cmidrule(lr){1-1}  \cmidrule(lr){2-12} \cmidrule(lr){13-14} 

                CLIP (\textit{Bs.})~\cite{radford2021learning}
                & ViT-B/16
                &$65.54$ 
                &$62.91$ 
                &$61.54$ 
                &$57.75$ 
                &$57.88$ 
                &$57.89$ 
                &$56.62$ 
                &$55.40$ 
                &$54.20$ 
                &$54.23$
                &$55.06$
                &$58.09$
                &$\textbf{10.48}$ \\
                \textbf{LP-DiF (\textit{Ours})} 
                & ViT-B/16
                &$\textbf{83.94}$ 
                &$\textbf{80.59}$ 
                &$\textbf{79.17}$ 
                &$\textbf{74.30}$ 
                &$\textbf{73.89}$
                &$\textbf{73.44}$ 
                &$\textbf{71.60}$ 
                &$\textbf{70.81}$ 
                &$\textbf{69.08}$ 
                &$\textbf{68.74}$ &$\textbf{68.53}$ &$\textbf{74.00}$ 
                &$15.41$\\
\midrule
\midrule
                {Joint-LP (\textit{UB. of ours})} 
                & {ViT-B/16}
                &${{83.94}}$ 
                &${{80.83}}$ 
                &${{79.43}}$
                &${{77.06}}$
                &${{76.35}}$
                &${{74.89}}$
                &${{73.66}}$
                &${{72.79}}$
                &${{71.84}}$
                &${{72.06}}$
                &${{71.88}}$
                &${{75.88}}$
                &${{12.06}}$ \\
    
				\bottomrule
			\end{tabular}
		\label{table:cub-200}
	\end{center}
\end{table*}

\begin{table}[t]
\centering
		\caption{\textbf{Comparison} between our LP-DiF and other replay-based FSCIL SOTA methods on \textit{mini}-ImageNet. ``\textbf{Exemplar / cls}'' represents the number of exemplar of each class. Note that these existing  FSCIL SOTA methods use the units of image as exemplar. ``\textbf{Disk Space / cls}'' represents the disk space consumed by the exemplar of each class.}
 \fontsize{8}{10}\selectfont
 \setlength{\tabcolsep}{8.5pt}
			\begin{tabular}{l|ccc}
				\toprule
				\textbf{Methods}
                &\textbf{Exemplar / cls}
                &\textbf{Disk Space / cls}
                &\textbf{Avg}  \\ 

\cmidrule(lr){1-1}  \cmidrule(lr){2-2} \cmidrule(lr){3-3} \cmidrule(lr){4-4} 
                Replay~\cite{liu2022few}
               &{1 image}
                &51.19 KB
                &58.02 \\  
                
                F2M~\cite{shi2021overcoming}
                &{5 images}
                &255.95 KB
                &57.89 \\

                BDF~\cite{zhao2023few}
                &{1 image}
                &51.19 KB
                &61.42 \\

                \textbf{LP-DiF} 
                & \textbf{2 vectors}
                &$\textbf{1.15}$ \textbf{KB}
                &$\textbf{93.76}$ \\

				\bottomrule
			\end{tabular}
		\label{table:replay_sota}

\end{table}

\begin{table*}[t] 
	
	\begin{center}
		\caption{\textbf{Comparison} with state-of-the-art FSCIL methods on {CIFAR-100}. ``\textbf{Backbone.}'' represents the backbone of visual model.``\textbf{Avg}'' represents the average accuracy of all sessions; the higher the value, the better performance. ``\textbf{PD}'' represents the Performance Drop rate; the lower the value, the better performance. ``\textbf{bs}'' is the abbreviation of ``baseline''.  ``\textbf{\textit{UB.}}'' is the abbreviation of ``Upper bound''.}
		
     \fontsize{8}{10}\selectfont
 \setlength{\tabcolsep}{7.3pt}
			\begin{tabular}{lccccccccccccc}
				\toprule
				\multicolumn{1}{l}{\multirow{2}{*}{\textbf{Methods}}} & \multicolumn{1}{c}{\multirow{2}{*}{\textbf{Backbone}}} &\multicolumn{9}{c}{\textbf{Accuracy in each session (\%)} $\uparrow$} & \multicolumn{1}{l}{\multirow{2}{*}{\textbf{Avg} $\uparrow$}} & \multicolumn{1}{l}{\multirow{2}{*}{\textbf{PD} $\downarrow $}}  \\ 
				\cmidrule{3-11}
				& &  0   &  1      &  2      &  3    &  4     &  5  & \ 6     &  7      & 8   &  &  \\ 
\midrule
                
				TOPIC~\cite{tao2020few}  
                & Res18
                &$64.10$	
                &$55.88$	
                &$47.07$	
                &$45.16$	
                &$40.11$	
                &$36.38$	
                &$33.96$	
                &$31.55$	
                &$29.37$	
                &$42.62$ 
                &$34.73$\\
    
                
                F2M~\cite{shi2021overcoming} 
                & Res18
                &$64.71$ 
                &$62.05$ 
                &$59.01$ 
                &$55.58$ 
                &$52.55$ 
                &$49.96$ 
                &$48.08$
                &$46.28$ 
                &$44.67$ 
                &$53.65$ 
                &$20.04$ \\

				CEC~\cite{zhang2021few}
                & Res18
                &$73.07$	
                &$68.88$
                &$65.26$	
                &$61.19$	
                &$58.09$	
                &$55.57$	
                &$53.22$	
                &$51.34$	
                &$49.14$	
                &$59.53$
                &$23.93$\\

				Replay~\cite{liu2022few} 
                & Res18
                &$74.40$	
                &$70.20$	
                &$66.54$	
                &$62.51$	
                &$59.71$	
                &$56.58$	
                &$54.52$	
                &$52.39$	
                &$50.14$	
                &$60.78$ 
                &$24.26$\\
                
				MetaFSCIL~\cite{chi2022metafscil} 
                & Res18
                &$74.50$
                &$70.10$
                &$66.84$
                &$62.77$
                &$59.48$
                &$56.52$
                &$54.36$
                &$52.56$
                &$49.97$
                &$60.79$
                &$24.53$ \\

                GKEAL~\cite{zhuang2023gkeal} 
                & Res18
                &$74.01$
                &$70.45$ 
                &$67.01$ 
                &$63.08$ 
                &$60.01$ 
                &$57.30$ 
                &$55.50$ 
                &$53.39$ 
                &$51.40$ 
                &$61.35$ 
                &$22.61$ \\

                C-FSCIL~\cite{hersche2022constrained}
                & Res12
                &$77.47$
                &$72.40$ 
                &$67.47$ 
                &$63.25$ 
                &$59.84$ 
                &$56.95$ 
                &$54.42$ 
                &$52.47$ 
                &$50.47$ 
                &$61.64$ 
                &$27.00$ \\

                FCIL~\cite{gu2023few}
                & Res18
                &$77.12$ 
                &$72.42$ 
                &$68.31$ 
                &$64.47$ 
                &$61.18$ 
                &$58.17$ 
                &$56.06$ 
                &$54.19$ 
                &$52.02$ 
                &$62.66$ 
                &$25.10$ \\

                FACT~\cite{zhou2022forward}
                & Res18
                &$78.22$ 
                &$72.40$ 
                &$68.57$ 
                &$64.73$ 
                &$61.40$ 
                &$58.57$ 
                &$56.30$ 
                &$53.83$ 
                &$51.72$ 
                &$62.86$ 
                &$26.50$ \\
                
                SAVC~\cite{song2023learning}
                & Res18
                &$78.77$ 
                &$73.31$ 
                &$69.31$ 
                &$64.93$ 
                &$61.70$ 
                &$59.25$ 
                &$57.13$ 
                &$55.19$ 
                &$53.12$ 
                &$63.63$ 
                &$25.65$\\

                BiDistFSCIL~\cite{zhao2023few}
                & Res18
                &$79.45$ 
                &$75.38$ 
                &$71.84$ 
                &$67.95$ 
                &$64.96$ 
                &$61.95$ 
                &$60.16$ 
                &$57.67$ 
                &$55.88$ 
                &$66.14$ 
                &$23.57$ \\
                
                NC-FSCIL~\cite{yang2022neural}
                & Res18
                &$\textbf{82.52}$
                &$76.82$
                &$73.34$
                &$69.68$
                &$66.19$	
                &$62.85$
                &$60.96$
                &$59.02$
                &$56.11$
                &$67.50$
                &$26.41$ \\
   \cmidrule(lr){1-1}  \cmidrule(lr){2-10} \cmidrule(lr){11-12} 

                CLIP (\textit{Bs.})~\cite{radford2021learning} 
                & ViT-B/16
                &$74.44$ 
                &$72.96$ 
                &$72.21$ 
                &$70.49$ 
                &$70.18$ 
                &$70.00$ 
                &$69.81$ 
                &$69.23$ 
                &$68.37$ 
                &$70.86$
                &$\textbf{6.07}$ \\
                \textbf{LP-DiF (\textit{Ours})} 
                & ViT-B/16
                &${80.23}$ 
                &$\textbf{77.75}$ 
                &$\textbf{76.78}$ 
                &$\textbf{74.62}$ 
                &$\textbf{74.03}$
                &$\textbf{73.87}$ 
                &$\textbf{73.84}$ 
                &$\textbf{72.96}$ 
                &$\textbf{72.02}$ 
                &$\textbf{75.12}$ 
                &$8.21$\\

\midrule
\midrule

                {Joint-LP (\textit{UB. of ours})}
                & {ViT-B/16}
                &${{80.23}}$ 
                &${{79.85}}$ 
                &${{78.63}}$
                &${{76.13}}$
                &${{75.31}}$
                &${{74.67}}$
                &${{74.24}}$
                &${{73.58}}$
                &${{73.35}}$
                &${{76.22}}$
                &${{6.88}}$ \\
    
				\bottomrule
			\end{tabular}
		\label{table:cifar-100}
	\end{center}

\end{table*}

\begin{figure}[t]

    \includegraphics[width=1\linewidth]{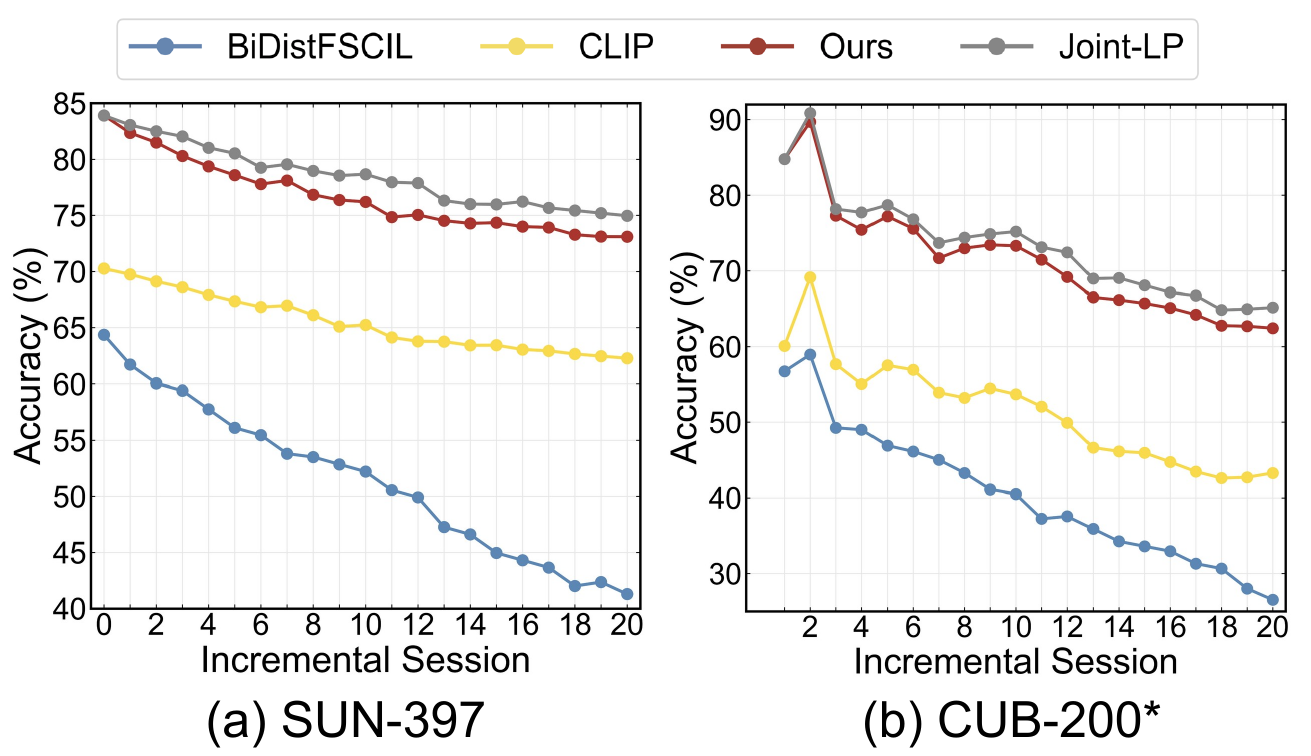}

    \caption{\textbf{Accuracy curves} of our LP-DiF and comparison with counterparts on {(a) SUN-397} and {(b) CUB200*}. our LP-DiF method significantly surpasses both CLIP and BiDistFSCIL, and attains performance levels that are very close to the respective upper bounds.} 
    \label{fig:more}
    
    \end{figure}

\begin{table}[t]
\centering
		\caption{Comparison with \textbf{standard CIL methods based on pre-trained models} on the three common benchmarks in terms of \textbf{Avg}. $^\ddag$ indicates our reproduction on FSCIL protocol. }
        \fontsize{9}{12}\selectfont
        \setlength{\tabcolsep}{4.3pt}
			\begin{tabular}{l|ccc}
				\toprule
				\textbf{Methods}
                &\textbf{CIFAR-100}
                &\textbf{\textit{mini}-ImageNet}
                &\textbf{CUB-200}  \\ 


\cmidrule(lr){1-1}  \cmidrule(lr){2-2} \cmidrule(lr){3-3} \cmidrule(lr){4-4} 

                L2P$^\ddag$~\cite{wang2022learning}
                &$61.77$
                &$75.68$
                &$56.95$  \\

                DualPrompt$^\ddag$~\cite{wang2022dualprompt}
                &$63.50$
                &$76.61$
                &$62.32$  \\

                AttriCLIP$^\ddag$~\cite{wang2023attriclip}
                &$59.24$
                &$81.74$
                &$47.81$  \\

 \cmidrule(lr){1-1}  \cmidrule(lr){2-2} \cmidrule(lr){3-3} \cmidrule(lr){4-4}                
                \textbf{LP-DiF (\textit{Ours})}
                &$\textbf{75.12}$
                &$\textbf{93.76}$
                &$\textbf{74.00}$  \\

				\bottomrule
			\end{tabular}
		\label{table: other_pretrained}

\end{table}

\begin{table}[t]
\centering
		\caption{\textbf{Comparison} with  SOTA FSCIL methods combined with CLIP on three common benchmarks in terms of \textbf{Avg}.  We replace the original backbone of these methods with CLIP's image encoder (ViT-B/16), and we use the text encoding generated by CLIP's text encoder for each category as initialization for the classification layer parameters of these methods.}
        \fontsize{8.8}{12}\selectfont
        \setlength{\tabcolsep}{2pt}
			\begin{tabular}{l|ccc}
				\toprule
				\textbf{Methods}
                &\textbf{CIFAR-100}
                &\textbf{\textit{mini}-ImageNet}
                &\textbf{CUB-200}  \\ 

\cmidrule(lr){1-1}  \cmidrule(lr){2-2} \cmidrule(lr){3-3} \cmidrule(lr){4-4} 
                
                SAVC~\cite{song2023learning} + CLIP
                &$68.46$
                &$86.81$
                &$71.66$ \\

                BiDistFSCIL~\cite{zhao2023few} + CLIP
                &$69.40$
                &$84.63$
                &$70.95$ \\

                \textbf{LP-DiF (\textit{Ours})} 
                &$\textbf{75.12}$
                &$\textbf{93.76}$
                &$\textbf{74.00}$  \\

				\bottomrule
			\end{tabular}
		\label{table:combine_clip}

\end{table}

\subsection{Main Results}
\noindent\textbf{Comparison with State-of-The-Arts.} We summarize the results of competing methods on \textit{mini}-ImageNet in Table~\ref{table:imgnet}.
Clearly, employing CLIP (baseline)~\cite{thengane2022clip,radford2021learning} for zero-shot evaluation alone outperforms all existing FSCIL methods by a large margin in terms of accuracy in each session and Average Accuracy (Avg). Naturally, it achieves a notably lower Performance Drop rate (PD).
%
%
Our LP-DiF further achieves \textbf{16.19\%} ($77.57$\% $\rightarrow$ \textbf{$93.76$\%}) gains than the CLIP in terms of Avg, and shows comparable PD performance, \ie, \textbf{$4.66$\%} \vs \textbf{$4.56$\%}.
As for the existing SOTA methods, \eg,  NC-FSCIL~\cite{yang2022neural}, which presents the best Avg among all the SOTA methods, LP-DiF gains \textbf{25.94\%} improvements, \ie, 67.82\% $\rightarrow$ \textbf{93.76\%}.
Comparing with C-FSCIL~\cite{hersche2022constrained}, which presents the best PD. among the competing methods, LP-DiF gains \textbf{10.33\%} improvements, \ie, 14.99\% $\rightarrow$ \textbf{4.66\%}.
 %
%
%
%
Tab.~\ref{table:replay_sota} highlights the comparison with the replay-based FSCIL method. 
Note that existing replay-based methods directly store old-class images, while our method only requires storing one mean vector and one variance vector (diagonal elements of the covariance matrix) for each old class.
Generally, LP-DiF significantly outperforms others in terms of performance while costing the least amount of storage space.
Tab.~\ref{table:cub-200} and Tab.~\ref{table:cifar-100} shows the comparison results on {CUB-200} and {CIFAR-100}.
%
Overall, the performance of our LP-DiF can be summarized in two points.
\textbf{1)} There are significant improvements compared to CLIP (\textit{baseline}) in terms of Avg (\ie, \textbf{15.91\%} and \textbf{4.26\%} improvements on CUB-200 and CIFAR-100 respectively). 
\textbf{2)} Compared to existing SOTA methods, LP-DiF achieves a higher Avg and lower PD.
%
%
Moreover, considering that our method use CLIP as backbone, which is more stronger than those existing FSCIL methods whose backbone are ResNet, we replaced the backbone of these methods with CLIP to make a fairer comparison.
Specifically, we replace the original backbone of these methods with CLIP's image encoder (ViT-B/16), and we use the text encoding generated by CLIP's text encoder for each category as initialization for the classification layer parameters of these methods.
Tab.~\ref{table:combine_clip} shows that the performance of the existing SOTA FSCIL methods combined with CLIP are still lower than our LP-DiF.
These above results clearly illustrate the superiority of our LP-DiF.

\vspace{5pt}
\noindent\textbf{Comparison with Upper Bound.}
Assuming that the training set from each previous session is available, we can jointly train the prompts using these sets, thereby avoiding the issue of forgetting old information.
In class-incremental learning, the above-mentioned setting can be considered as an upper bound, and serve as a reference for evaluating FSCIL method.
Thus, we compare our LP-DiF with its upper bound
(\ie \textbf{Joint-LP}).
As shown in the last row of Tab.~\ref{table:imgnet}, Tab.~\ref{table:cub-200} and Tab.~\ref{table:cifar-100} across the three benchmarks, the performances of our method are very close to the upper bounds in terms of Avg, with the largest gap being only $1.05\%$, $1.42\%$ and $1.10\%$ on \textit{mini}-ImageNet, {CUB-200} and CIFAR-100, respectively.
The results indicate that our LP-DiF is highly effective in preventing catastrophic forgetting. 
It is noted that NC-FSCIL achieves higher accuracy than both ours and Joint-LP.
The architectures of NC-FSCIL, which is ResNet-based method, and Joint-LP, which is CLIP-based method, are different; NC-FSCIL trains all layers of the model during the base session, whereas Joint-LP only train the prompt. Therefore, it is acceptable that NC-FSCIL outperforms Joint-LP in session 0.

\vspace{5pt}
\noindent\textbf{Comparison with Pre-trained Models-based Standard CIL Methods.}
To further demonstrate the superiority of our method, we compare it with several recent standard CIL~\cite{zhou2023deep} methods which also utilize pre-trained models: L2P~\cite{wang2022learning}, DualPrompt~\cite{wang2022dualprompt} and AttriCLIP~\cite{wang2023attriclip}.
The L2P and DualPrompt are based on a pretrained ViT and learn the visual prompts to solve CIL problems, while AttriCLIP builds on CLIP and training different text prompts to encode different knowledge.
We reproduce these three approaches on CIFAR-100, CUB-200, and \textit{mini}-ImageNet and evaluate them under FSCIL protocol respectively.
As shown in Tab.~\ref{table: other_pretrained}, our LP-DiF outperforms these methods by a large margin in terms of Avg across all three benchmarks.
We also find that these methods based on pre-trained models underperform BiDistFSCIL~\cite{zhao2023few} on CIFAR-100 and CUB-200.
This indicates that these methods are not advantageous for the FSCIL setting, further underscoring the effectiveness and significance of our method.

\vspace{5pt}
\noindent\textbf{More Challenging Benchmarks.} 
On the three widely used benchmarks, the performance of our LP-DiF closely approaches the upper bound.
To further assess our LP-DiF, we provide two more challenging benchmarks: {SUN-397} and {CUB-200*}.
For each challenging benchmark, we compare our LP-DiF with three distinct approaches: zero-shot evaluation using CLIP (baseline), Joint-LP (upper bound), and BiDistFSCIL~\cite{zhao2023few} (SOTA open-source method). The corresponding performance curves are depicted in Fig.~\ref{fig:more}.
Overall, on both {SUN-397} and {CUB-200*}, our LP-DiF method \textbf{1)} significantly surpasses both CLIP and BiDistFSCIL, and \textbf{2)} attains performance levels that are very close to the respective upper bounds.
The results show that our LP-DiF remains very effective on these challenging situations, including those with a larger number of classes and extended session lengths, \eg, $397$ classes across $21$ sessions in {SUN-397}, as well as in those without a base session, exemplified by {CUB-200*}.

\begin{table*}[t] 
	\begin{center}
		\centering
		\caption{\textbf{Ablation studies} of our LP-DiF on {\textit{mini}-ImageNet}. \textbf{\texttt{LP}} and \textbf{\texttt{OCD}} denote learning prompts and old-class distribution, respectively. \textbf{\texttt{RF}} and \textbf{\texttt{SF}} denote the real features of training images and synthesized features generated by VAE, respectively.}
  
   \fontsize{8}{10}\selectfont
 \setlength{\tabcolsep}{8pt}
			\begin{tabular}{cccccccccccccccc}
				\toprule
				\multicolumn{1}{c}{\multirow{2}{*}{\textbf{\texttt{CLIP}}}} & \multicolumn{1}{c}{\multirow{2}{*}{\textbf{\texttt{LP}}}} &\multicolumn{2}{c}{\textbf{\texttt{OCD}}} &\multicolumn{9}{c}{\textbf{Accuracy in each session (\%)} $\uparrow$} & \multicolumn{1}{l}{\multirow{2}{*}{\textbf{Avg} $\uparrow$}} & \multicolumn{1}{l}{\multirow{2}{*}{\textbf{PD} $\downarrow $}}  \\ 
    
				\cmidrule{5-13}
				& &\textbf{\texttt{RF}}& \textbf{\texttt{SF}} &0&1&2&3&4&5&6&7&8& &  \\ 
\midrule
                
                \checkmark
                &
                &
                &
                &$80.01$ 
                &$79.16$ 
                &$78.89$ 
                &$77.97$ 
                &$77.44$ 
                &$76.83$ 
                &$76.32$ 
                &$76.02$ 
                &$75.45$ 
                &$77.57$
                &$\textbf{4.56}$ \\

                \checkmark
                &\checkmark
                &
                &
                &${\textbf{96.34}}$ 
                &${94.28}$ 
                &${92.83}$ 
                &${89.93}$ 
                &${88.39}$
                &${86.10}$ 
                &${85.49}$ 
                &${85.70}$ 
                &${84.76}$ 
                &${89.31}$ 
                &$11.58$\\

                \checkmark
                &\checkmark
                &\checkmark
                &
                &${\textbf{96.34}}$ 
                &${\textbf{96.14}}$ 
                &${94.01}$ 
                &${94.27}$ 
                &${93.23}$
                &${93.07}$ 
                &${91.34}$ 
                &${91.17}$ 
                &${90.76}$ 
                &${93.37}$ 
                &$5.46$\\

                \checkmark
                &\checkmark
                &
                &\checkmark
                &${\textbf{96.34}}$ 
                &${\textbf{96.14}}$ 
                &${93.79}$ 
                &${92.48}$ 
                &${91.25}$
                &${90.94}$ 
                &${90.15}$ 
                &${89.41}$ 
                &${89.27}$ 
                &${92.23}$ 
                &$7.07$\\

                \checkmark
                &\checkmark
                &\checkmark
                &\checkmark
                &$\textbf{\textbf{96.34}}$ 
                &$\textbf{96.14}$ 
                &$\textbf{94.62}$ 
                &$\textbf{94.37}$ 
                &$\textbf{94.06}$
                &$\textbf{93.44}$ 
                &$\textbf{92.21}$ 
                &$\textbf{92.29}$ 
                &$\textbf{91.68}$ 
                &$\textbf{93.76}$ 
                &$4.66$\\
				\bottomrule
			\end{tabular}
		\label{table:ablation_mini}
	\end{center}
\end{table*}

\begin{table*}[t] 
	\begin{center}
		\centering
		\caption{\textbf{Ablation studies} of our LP-DiF on {CUB-200}. \textbf{\texttt{LP}} and \textbf{\texttt{OCD}} denote learning prompts and old-class distribution, respectively. \textbf{\texttt{RF}} and \textbf{\texttt{SF}} denote the real features of training images and synthesized features generated by VAE, respectively.}
  
   \fontsize{8}{10}\selectfont
 \setlength{\tabcolsep}{6pt}
			\begin{tabular}{cccccccccccccccccc}
				\toprule
				\multicolumn{1}{c}{\multirow{2}{*}{\textbf{\texttt{CLIP}}}} & \multicolumn{1}{c}{\multirow{2}{*}{\textbf{\texttt{LP}}}} &\multicolumn{2}{c}{\textbf{\texttt{OCD}}} &\multicolumn{11}{c}{\textbf{Accuracy in each session (\%)} $\uparrow$} & \multicolumn{1}{l}{\multirow{2}{*}{\textbf{Avg} $\uparrow$}} & \multicolumn{1}{l}{\multirow{2}{*}{\textbf{PD} $\downarrow $}}  \\ 
    
				\cmidrule{5-15}
				& &\textbf{\texttt{RF}}& \textbf{\texttt{SF}} &0&1&2&3&4&5&6&7&8&9&10 & &  \\ 
\midrule
                
                \checkmark
                &
                &
                &
                &$65.54$ 
                &$62.91$ 
                &$61.54$ 
                &$57.75$ 
                &$57.88$ 
                &$57.89$ 
                &$56.62$ 
                &$55.40$ 
                &$54.20$ 
                &$54.23$
                &$55.06$
                &$58.09$
                &$\textbf{10.48}$ \\

                \checkmark
                &\checkmark
                &
                &
                &$\textbf{83.94}$ 
                &$78.32$ 
                &$75.10$ 
                &$70.62$ 
                &$70.75$ 
                &$68.09$ 
                &$65.69$ 
                &$64.55$ 
                &$62.47$ 
                &$61.94$
                &$61.96$
                &$70.71$
                &$21.98$ \\

                \checkmark
                &\checkmark
                &\checkmark
                &
                &$\textbf{83.94}$ 
                &$\textbf{80.59}$ 
                &$78.83$ 
                &$73.66$ 
                &$73.24$ 
                &$72.54$ 
                &$70.57$ 
                &$69.72$ 
                &$68.88$ 
                &$67.86$
                &$67.90$
                &$73.43$
                &$16.04$ \\

                \checkmark
                &\checkmark
                &
                &\checkmark
                &$\textbf{83.94}$ 
                &$\textbf{80.59}$ 
                &$78.41$ 
                &$72.65$ 
                &$72.76$ 
                &$71.25$ 
                &$69.86$ 
                &$67.99$ 
                &$67.20$ 
                &$66.73$
                &$66.88$
                &$72.56$
                &$17.06$ \\

                \checkmark
                &\checkmark
                &\checkmark
                &\checkmark
                &$\textbf{83.94}$ 
                &$\textbf{80.59}$ 
                &$\textbf{79.17}$ 
                &$\textbf{74.30}$ 
                &$\textbf{73.89}$
                &$\textbf{73.44}$ 
                &$\textbf{71.60}$ 
                &$\textbf{70.81}$ 
                &$\textbf{69.08}$ 
                &$\textbf{68.74}$ &$\textbf{68.53}$ &$\textbf{74.00}$ 
                &$15.41$\\
				\bottomrule
			\end{tabular}
		\label{table:ablation_cub}
	\end{center}
\end{table*}

\begin{table*}[t] 
	\begin{center}
		\centering
		\caption{\textbf{Ablation studies} of our LP-DiF on {CIFAR-100}. \textbf{\texttt{LP}} and \textbf{\texttt{OCD}} denote learning prompts and old-class distribution, respectively. \textbf{\texttt{RF}} and \textbf{\texttt{SF}} denote the real features of training images and synthesized features generated by VAE, respectively.}
  
   \fontsize{8}{10}\selectfont
 \setlength{\tabcolsep}{8pt}
			\begin{tabular}{cccccccccccccccc}
				\toprule
				\multicolumn{1}{c}{\multirow{2}{*}{\textbf{\texttt{CLIP}}}} & \multicolumn{1}{c}{\multirow{2}{*}{\textbf{\texttt{LP}}}} &\multicolumn{2}{c}{\textbf{\texttt{OCD}}} &\multicolumn{9}{c}{\textbf{Accuracy in each session (\%)} $\uparrow$} & \multicolumn{1}{l}{\multirow{2}{*}{\textbf{Avg} $\uparrow$}} & \multicolumn{1}{l}{\multirow{2}{*}{\textbf{PD} $\downarrow $}}  \\ 
    
				\cmidrule{5-13}
				& &\textbf{\texttt{RF}}& \textbf{\texttt{SF}} &0&1&2&3&4&5&6&7&8& &  \\ 
\midrule
                
                \checkmark
                &
                &
                &
                &$74.44$ 
                &$72.96$ 
                &$72.21$ 
                &$70.49$ 
                &$70.18$ 
                &$70.00$ 
                &$69.81$ 
                &$69.23$ 
                &$68.37$ 
                &$70.86$
                &$\textbf{6.07}$ \\

                \checkmark
                &\checkmark
                &
                &
                &${\textbf{80.23}}$ 
                &${75.81}$ 
                &${75.03}$ 
                &${71.65}$ 
                &${71.67}$
                &${70.94}$ 
                &${70.48}$ 
                &${70.01}$ 
                &${69.54}$ 
                &${72.81}$ 
                &$10.69$\\

                \checkmark
                &\checkmark
                &\checkmark
                &
               &${\textbf{80.23}}$ 
                &${\textbf{77.75}}$ 
                &${\textbf{76.84}}$ 
                &${74.40}$ 
                &${73.81}$
                &${73.24}$ 
                &${73.69}$ 
                &${72.52}$ 
                &${71.60}$ 
                &${74.89}$ 
                &$8.63$\\

                \checkmark
                &\checkmark
                &
                &\checkmark
               &${\textbf{80.23}}$ 
                &${\textbf{77.75}}$ 
                &${75.63}$ 
                &${73.75}$ 
                &${73.09}$
                &${72.36}$ 
                &${72.31}$ 
                &${71.84}$ 
                &${70.76}$ 
                &${74.19}$ 
                &$9.47$\\

                \checkmark
                &\checkmark
                &\checkmark
                &\checkmark
                &$\textbf{\textbf{80.23}}$ 
                &$\textbf{77.75}$ 
                &$76.78$ 
                &$\textbf{74.62}$ 
                &$\textbf{74.03}$
                &$\textbf{73.87}$ 
                &$\textbf{73.84}$ 
                &$\textbf{72.96}$ 
                &$\textbf{72.02}$ 
                &$\textbf{75.12}$ 
                &$8.21$\\
				\bottomrule
			\end{tabular}
		\label{table:ablation_cifar}
	\end{center}
\end{table*}

\begin{figure*}[t]
    \begin{center}
    \includegraphics[width=1.0\linewidth]{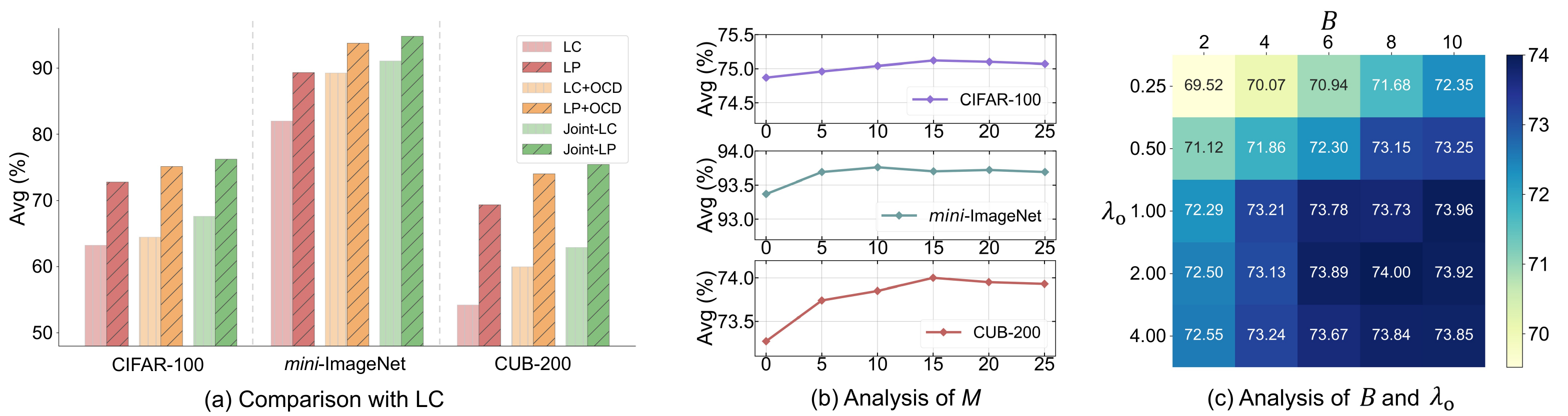}
    \end{center}
    \caption{\textbf{Ablation studies} of our LP-DiF. \textbf{(a)} Comparison with the method of incorporating a Linear Classifier (LC) into a pre-trained image encoder for training on three common benchmarks. \textbf{(b)} Analysis of $M$ on three common benchmarks. \textbf{(c)} Analysis of $B$ and $\lambda_\text{o}$ in terms of Avg on \textbf{CUB-200}.} 
    \label{fig:abl}
    \end{figure*}

\begin{figure}[t]
    \begin{center}
    \includegraphics[width=1\columnwidth]{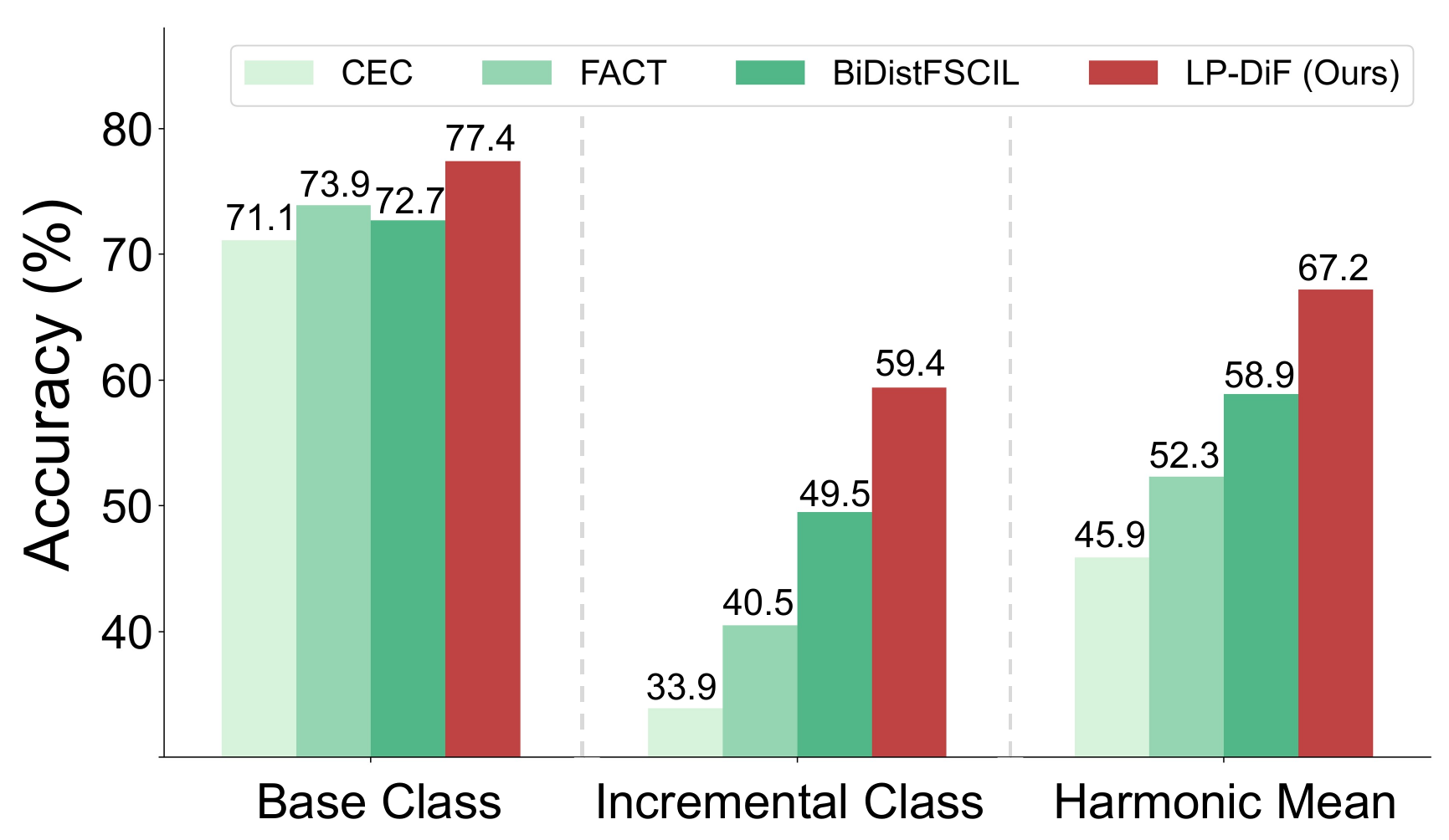}
    \end{center}
    \caption{\textbf{Decomposing} the performance of the \textbf{base class} and the \textbf{incremental class}. Their harmonic mean are also reported. The performance is evaluated by the model from the last session on \textbf{CUB-200}.}
    \label{fig:trade}
\end{figure}

\subsection{Ablation Studies and Analysis}\label{sec:ablation}

\noindent\textbf{Analysis of Key Components.}
Our proposed method involves prompt tuning on CLIP to adapt the knowledge from each incremental session. It also constructs feature-level distributions to preserve old knowledge, thereby achieving resistance to catastrophic forgetting.
To investigate the effect of the key components in our method, \ie, CLIP, prompt learning ({LP}), the distribution estimated by real features (RF) of training images, and the synthesized features (SF), we summarized the performance of each component on three common FSCIL benchmark in Tab.~\ref{table:ablation_mini}, Tab.~\ref{table:ablation_cub} and Tab.~\ref{table:ablation_cifar}.
%
Take the results on \textit{mini}-ImageNet as an example, as illustrated in Tab.~\ref{table:ablation_mini}, employing the {LP} technique noticeably improves performance across each session, ultimately resulting in a superior \textbf{$11.74$\%} performance, \ie, from $77.57$\% $\rightarrow$ \textbf{$89.31$\%} in terms of average performance (refer to the second row of Table~\ref{table:ablation_mini}).
However, solely implementing {LP} causes higher PD than CLIP, \eg, \textbf{$4.56$\%} $\rightarrow$ $11.58$\%, due to the forgetting of old knowledge during learning in new sessions.
Additionally, as mentioned in Sec.\ref{sec:construct}, the old-class distribution (OCD) is effective in tackling the forgetting problem.
Note that the distribution of each incremental session is estimated by real features (RF) and the synthesized features (SF) generated by VAE.
So we conducted separate evaluations to assess the effect of these two types of ``features''.
Concretely, using only RF to estimate the old-class distribution can improve the Avg by \textbf{$4.11$\%}, \ie, $89.31$\% $\rightarrow$ \textbf{93.42\%}, and reduce the PD. by \textbf{6.12\%}, \ie, $11.58$\% $\rightarrow$ \textbf{5.46\%}, (see the third row of Table~\ref{table:ablation_mini}).
Using only SF for each incremental session can also improve Avg and reduce PD, however, its effectiveness is marginally inferior to using only RF (refer to the fourth row of Table~\ref{table:ablation_mini}).
Finally, using both types of ``features'' can further improve the performance, which surpasses the outcomes achieved by using either RF or SF alone (see the last row of Table~\ref{table:ablation_mini}).
Thus, although LP can enable the model to effectively capture the knowledge of each session and improve performance, it still leads to catastrophic forgetting, while OCD can effectively prevent this issue.

\vspace{5pt}
\noindent\textbf{Analysis of $M$.}
As mentioned in Sec.~\ref{sec:construct}, $M$ is the number of synthetic features which participate in estimating $\bm{\mu}_c$ and $\bm{\sigma}_c$ of old-class distribution.
The value of $M$ will affect the accuracy of the estimated distribution, thus influencing the performance of the model.
Therefore, we test the effect of $M$ on the performance.
Fig.~\ref{fig:abl} (b) shows the results on the three widely used benchmarks.
Clearly, when $M$ increases from $0$, the Avg gradually improves.
 %
%
Nonetheless, when $M$ continues to increase, the Avg decreases slightly, possibly ascribing to that too many synthesized features can cause the estimated distribution to overly skew towards the distribution of the synthesized features.
%
In summary, $M=10$ (on \textit{mini}-ImageNet) or $M=15$ (on CIFAR-$100$ and CUB-$200$) are the best choices for performance. 

\vspace{5pt}
\noindent\textbf{Analysis of $B$ and $\lambda_o$.}
%
Here we investigate the effect of the two hyper-parameters, \ie, $B$, the number of selected old classes involved in $\mathcal{L}_\text{o}$, and $\lambda_o$, the tradeoff coefficient in $\mathcal{L}_\text{o}$.
The results on \textbf{CUB-200} are shown with a mixed matrix of these two hyper-parameters in Fig~\ref{fig:abl} (c).
%
%
Obviously, keeping $\lambda_o$ fixed, as $B$ increases, the Avg improves gradually.
Keeping $B$ fixed and tuning $\lambda_o$ shows a similar tendency with the above setting.
It achieves the best Avg when $B=8$ and $\lambda_o=2$ on CUB-200.
Then, when the values of $B$ and $\lambda_o$ are too large, \eg $B=10$ and $\lambda_o=4$, there is a slight performance drop.
These may be because too small values of $B$ and $\lambda_o$ lead to insufficient representation of old knowledge, while too large values may cause the model to overly emphasize old knowledge.


\vspace{5pt}
\noindent\textbf{Learning Prompt \vs Linear Classifier.}
This study utilizes prompt tuning to tailor CLIP to the specific knowledge of each session.
Another straightforward and intuitive strategy involves incorporating a linear classifier with the image encoder, which is initialized using the text encoding of handcrafted prompts.
So we conduct additional experiments: \textbf{1)} Refining the linear classifier (LC) solely with the training set accessible in the current session; \textbf{2)} Extending the first approach by integrating the old-class distribution for feature replay (LC + OCD); \textbf{3)} Jointly training the linear classifier with the complete training set from each session (Joint-LC).
%
%
As shown in Fig~\ref{fig:abl} (a), the Avg of LC is notably lower than that of LP across three wide benchmarks in terms of Avg.
The incorporation of OCD with LC (denoted as LC + OCD) enhances performance beyond LC alone, highlighting OCD's effectiveness in mitigating catastrophic forgetting.
Nevertheless, the combined LC + OCD is still inferior to LP + OCD.
In a joint training scenario, the performance of Joint-LC continues to be inferior to Joint-LP.
The results suggest that the strategy of learning prompts offers more merits for FSCIL than that of learning a linear classifier.

\begin{figure*}[t]
    \begin{center}
    \includegraphics[width=0.8\linewidth]{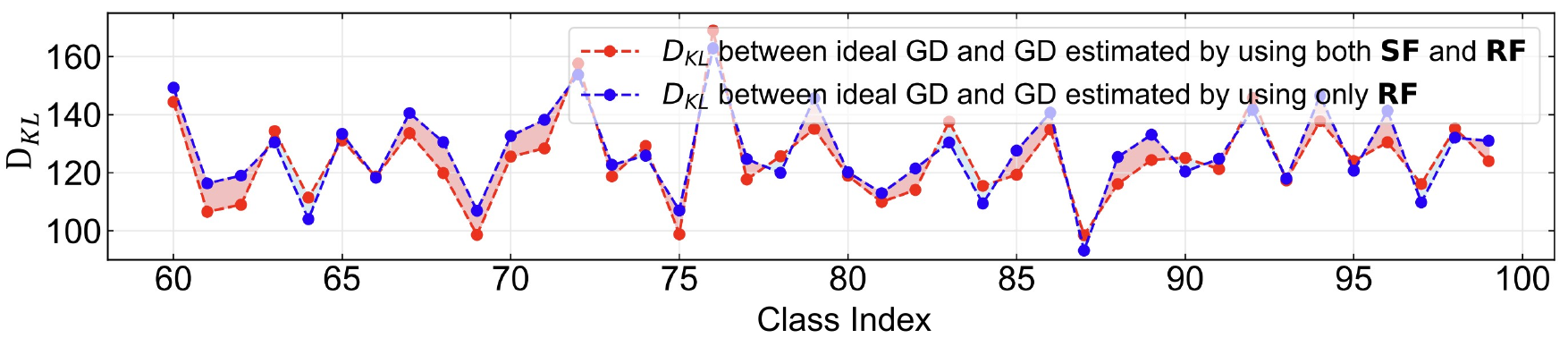}
    \end{center}
 \caption{\textbf{Analysis} the effectiveness of VAE on \textit{mini}-ImageNet. We calculate the difference between the estimated GD of each incremental class and corresponding reference GD (ideal GD, \ie, computed using the full training images of that class) by KL divergence. The red dots represents KL divergence between ideal GD and GD estimated by using both \textbf{SF} and \textbf{RF}. The blue dots represents KL divergence between ideal GD and GD estimated by using only \textbf{RF}.} 
    \label{fig:dis_diver}
\end{figure*}

\begin{table}[t] 
		\centering
		\caption{Comparison with other \textit{replay approaches} on {CUB-200} in terms of \textbf{Avg}. The results of ``Randomly selection'' are reports over $5$ runs with mean and standard deviations. iCaRL$^{\dagger}$ means applying the replay technique proposed in iCaRL to CLIP + LP.}
   \fontsize{8}{10}\selectfont
 \setlength{\tabcolsep}{8pt}
			\begin{tabular}{l|ccc}
				\toprule
				\textbf{Methods}
                &$\bm N_e$
                &\textbf{Disk Space}
                &\textbf{Avg}  \\ 
\cmidrule(lr){1-1}  \cmidrule(lr){2-2} \cmidrule(lr){3-3} \cmidrule(lr){4-4}

                CLIP + LP 
                &-
                &-
                &$69.36$  \\
\cmidrule(lr){1-1}  \cmidrule(lr){2-2} \cmidrule(lr){3-3} \cmidrule(lr){4-4}                 
                \multirow{4}{*}{Randomly selection}
                &1
                &$18.32 \pm 0.37$ MB
                &$69.95 \pm 0.56$  \\

                &2
                &$37.31 \pm 0.99$ MB
                &$71.16 \pm 0.24$  \\

                &3
                &$55.95 \pm 1.01$ MB
                &$72.44 \pm 0.09$  \\

                &4
                &$74.33 \pm 0.70$ MB
                &$73.64 \pm 0.16$  \\
\cmidrule(lr){1-1}  \cmidrule(lr){2-2} \cmidrule(lr){3-3} \cmidrule(lr){4-4}

                \multirow{4}{*}{iCaRL$^{\dagger}$~\cite{rebuffi2017icarl}}
                & 1
                &$18.54$ MB
                &$70.81$  \\

                & 2
                &$38.39$ MB
                &$71.68$  \\

                & 3
                &$55.40$ MB
                &$72.86$  \\

                & 4
                &$74.68$ MB
                &$73.95$  \\

%
\cmidrule(lr){1-1}  \cmidrule(lr){2-2} \cmidrule(lr){3-3} \cmidrule(lr){4-4} 
                \textbf{LP-DiF (\textit{Ours})} 
                &-
                &$\textbf{0.22}$ \textbf{MB}
                &$\textbf{74.00}$ \\
                
				\bottomrule
			\end{tabular}
		\label{table:replay}
\end{table}

\begin{figure}[t]
    \begin{center}
    \includegraphics[width=1\columnwidth]{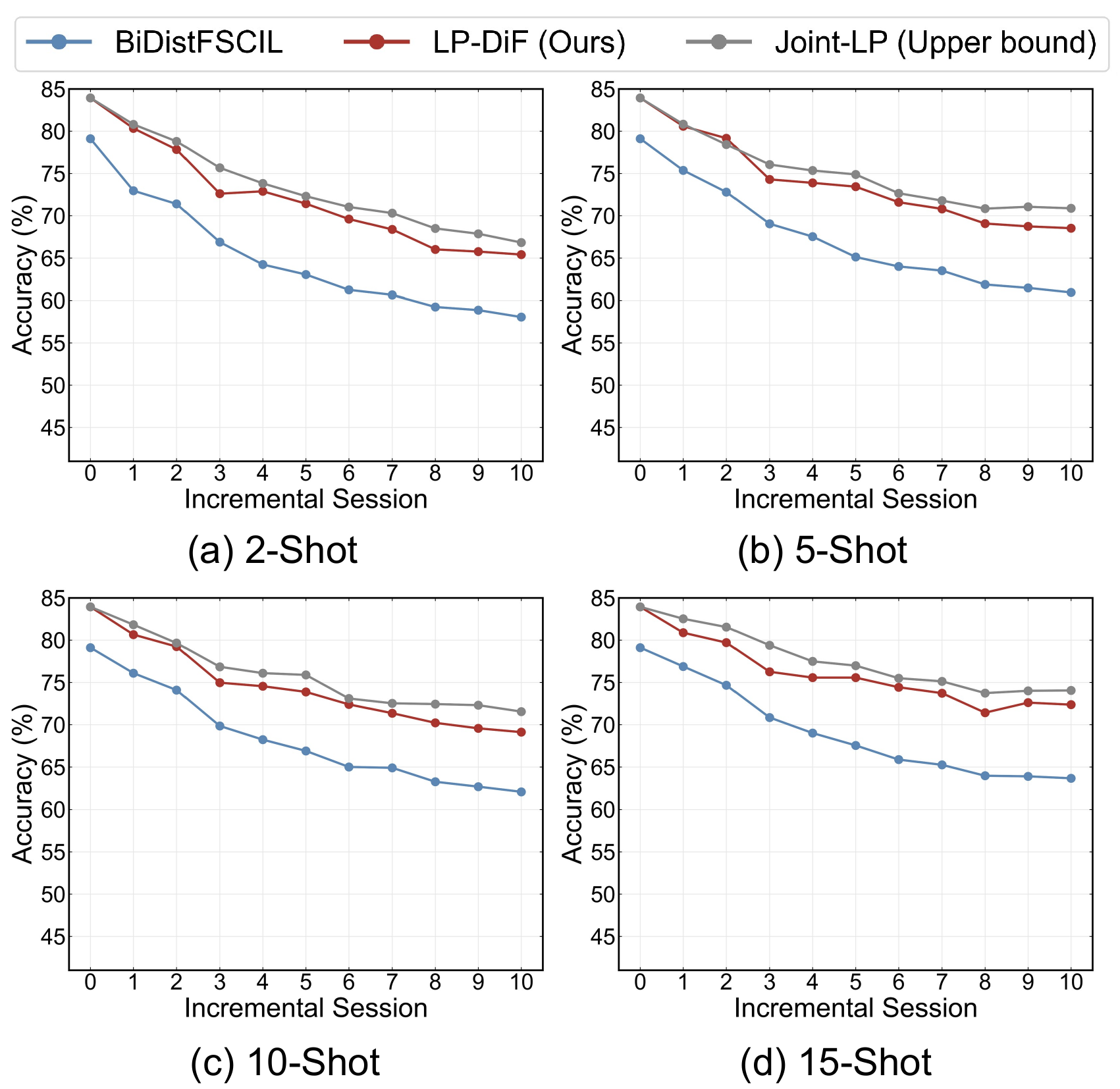}
    \end{center}
    \caption{Comparison with BiDistFSCIL (SOTA FSCIL method) and Joint-LP (Upper bound) under various \textbf{shot numbers} of incremental classes on CUB-200.}
    \label{fig:shot}
\end{figure}

\vspace{5pt}
\noindent\textbf{Old-Class Distribution \vs Image Exemplar.}
%
To further validate its efficacy in avoiding catastrophic forgetting, we compare our method with other replay-based approaches tailored for learning prompts, \ie, \textbf{1)} randomly selecting $N_e$ images of per old class as exemplars; \textbf{2)} adopting the replay strategy in iCaRL~\cite{rebuffi2017icarl}, specifically choosing $N_e$ images for each old class based on the proximity to the mean feature.
In addition, we execute the random selection approach five times, each with a different random seed, to reduce the uncertainty.
%
The average results with necessary storage space for replay on CUB-200 are shown in Table~\ref{table:replay}, where CLIP + LP indicates learning prompts sequentially across each incremental session without replay of old classes. (\eg the second row in Tab.~\ref{table:ablation_cub}).
Obviously, our method exhibits the best performance and lowest storage space in comparison to the two counterparts under various $N_e$. 
Especially, compared with iCaRL$^\dagger$ under $N_e =4$, LP-DiF shows a comparable performance while only requiring about $\textbf{0.002}\%$ storage space (thanks to the fact that we only store two vectors for each old class).
This underscores that our pseudo-feature replay technique can effectively combat catastrophic forgetting under conditions of light storage overhead.

\vspace{5pt}
\noindent\textbf{Decomposing the Performance of Base and Incremental Classes.}
Following previous studies~\cite{zhao2023few,zhang2021few,zhou2022forward}, in this section, we decompose the accuracy, respectively analyzing the effectiveness of our LP-DiF for the classes in the base session (\ie, base class) and for the classes in incremental sessions (\ie, incremental class), to evaluate if our method performs well on both base and incremental classes.
We report the comparison results in terms of individual accuracy of base and novel classes, as well as their harmonic mean, in the last session on CUB-200.
Fig.~\ref{fig:trade} shows that our LP-DiF outperforms the second best method on base class (\ie, FACT) by $3.5$\%, while outperforms the second best method on incremental class (\ie, BiDistFSCIL) by $9.9$\%.
Finally, the superior harmonic mean demonstrates our achievement of an enhanced balance between base and novel classes.

\vspace{5pt}
\noindent\textbf{Analysis on Shot Numbers.} 
To further demonstrate the superiority of our approach, we conducted experiments under various shot numbers of incremental classes.
Fig.~\ref{fig:shot} show the comparison results with BiDistFSCIL~\cite{zhao2023few} and Joint-LP on CUB-200 under \textbf{(a)} $2$-shot, \textbf{(b)} $5$-shot, \textbf{(c)} $10$-shot and \textbf{(d)} $15$-shot.
Obviously, across all the shot number settings, our LP-DiF consistently outperforms BiDistFSCIL significantly, and its performance is very close to the upper bound.
This result demonstrates that, regardless of the shot numbers of incremental classes, our LP-DiF presents satisfactory performance and the ability to resist catastrophic forgetting.

\vspace{5pt}
\noindent\textbf{Analysis on Effect of VAE.} 
From the results presented in Tab.~\ref{table:ablation_mini}, one can observe that using both synthesize features (\textbf{\texttt{SF}}) and real features (\textbf{\texttt{RF}}) for estimating the Gaussian distribution (GD), as compared to using only real features, can achieve higher performance. 
To elucidate the quality of the features synthesized by the VAE, we conducted the following analysis on \textit{mini}-ImageNet: 
we calculate the difference between the estimated GD of each incremental class and corresponding reference GD (ideal GD, \ie, computed using the full training images of that class) by \textit{KL divergence} $D_\text{KL}$
\begin{equation}
    D_\text{KL}(P|Q)=\frac{1}{2} \sum_{i=1}^{n} \left( \frac{\sigma_{p_i}^2}{\sigma_{q_i}^2} + \frac{(\mu_{q_i} - \mu_{p_i})^2}{\sigma_{q_i}^2} - 1 + \ln \left( \frac{\sigma_{q_i}^2}{\sigma_{p_i}^2} \right) \right),
\end{equation}
where $P$ and $Q$ represent the ideal GD and the estimated GD respectively, and $n$ represents the dimension of GD.
Statistically, lower $D_\text{KL}$ indicates that the estimated GD is closer to reference GD.
Fig~\ref{fig:dis_diver} shows the results on \textit{mini}-ImageNet.
Note that for most classes, the GD estimated using both \textbf{\texttt{SF}} and \textbf{\texttt{RF}} is closer to the reference distribution, indicating that \textbf{\texttt{SF}} can enrich more class-relevant information. 

\vspace{5pt}
\noindent\textbf{Training Time and Model Size.}
Compared to existing ResNet-based FSCIL methods, our LP-DiF is based on the heavier model (\ie, CLIP), which may raise concerns about model training efficiency and memory overhead. 
However, since LP-DiF only trains lightweight prompt vectors and a few layers of MLP in the VAE, it does not incur excessive computational costs.
Here, we offer some quantitative results for reference:
For the training time, with $8\times$ 2080ti GPUs, for each incremental session, training LP-DiF takes about 4.5 minutes (1.8 minutes for training VAE and 2.7 minutes for training prompts). 
In comparison, BiDistFSCIL~[{\color{green}5}] takes about 3.3 minutes for training for the same epochs.
For the volume  of trainable parameters, existed FSCIL methods relied on ResNet require training about 11.3M parameters for ResNet-18, respectively. 
However, LP-DiF only needs to train about 7.4M parameters for prompts and MLPs.
Thus, our LP-DiF achieved significant performance gain with acceptable addition on training time and lower volume of trainable parameters.



\section{Conclusion}
In this paper, we studied the FSCIL problem by introducing V-L pretrained model, and proposed \textbf{L}earning \textbf{P}rompt with \textbf{Di}stribution-based \textbf{F}eature replay (LP-DiF).
Specifically, prompt tuning is involved to adaptively capture the knowledge of each session.
To alleviate catastrophic forgetting, we established a feature-level distribution for each class, which is estimated by both real features of training images and synthesized features generated by a VAE decoder.
Then, pseudo features are sampled from old-class distributions, and combined with the training set of current session to train the prompts jointly.
Extensive experiments show that our LP-DiF achieves the new state-of-the-art in the FSCIL task.




%

\bibliographystyle{ieee_fullname}
\bibliography{reference}

\begin{thebibliography}{10}\itemsep=-1pt

\bibitem{agarwal2022semantics}
Aishwarya Agarwal, Biplab Banerjee, Fabio Cuzzolin, and Subhasis Chaudhuri.
\newblock Semantics-driven generative replay for few-shot class incremental learning.
\newblock In {\em Proceedings of the 30th ACM International Conference on Multimedia}, pages 5246--5254, 2022.

\bibitem{ahmad2022few}
Touqeer Ahmad, Akshay~Raj Dhamija, Steve Cruz, Ryan Rabinowitz, Chunchun Li, Mohsen Jafarzadeh, and Terrance~E Boult.
\newblock Few-shot class incremental learning leveraging self-supervised features.
\newblock In {\em Proceedings of the IEEE/CVF Conference on Computer Vision and Pattern Recognition}, pages 3900--3910, 2022.

\bibitem{ahmad2022variable}
Touqeer Ahmad, Akshay~Raj Dhamija, Mohsen Jafarzadeh, Steve Cruz, Ryan Rabinowitz, Chunchun Li, and Terrance~E Boult.
\newblock Variable few shot class incremental and open world learning.
\newblock In {\em Proceedings of the IEEE/CVF Conference on Computer Vision and Pattern Recognition}, pages 3688--3699, 2022.

\bibitem{akyurek2021subspace}
Afra~Feyza Aky{\"u}rek, Ekin Aky{\"u}rek, Derry~Tanti Wijaya, and Jacob Andreas.
\newblock Subspace regularizers for few-shot class incremental learning.
\newblock {\em arXiv preprint arXiv:2110.07059}, 2021.

\bibitem{aljundi2019gradient}
Rahaf Aljundi, Min Lin, Baptiste Goujaud, and Yoshua Bengio.
\newblock Gradient based sample selection for online continual learning.
\newblock {\em Advances in neural information processing systems}, 32, 2019.

\bibitem{bang2021rainbow}
Jihwan Bang, Heesu Kim, YoungJoon Yoo, Jung-Woo Ha, and Jonghyun Choi.
\newblock Rainbow memory: Continual learning with a memory of diverse samples.
\newblock In {\em Proceedings of the IEEE/CVF conference on computer vision and pattern recognition}, pages 8218--8227, 2021.

\bibitem{chaudhry2018riemannian}
Arslan Chaudhry, Puneet~K Dokania, Thalaiyasingam Ajanthan, and Philip~HS Torr.
\newblock Riemannian walk for incremental learning: Understanding forgetting and intransigence.
\newblock In {\em Proceedings of the European conference on computer vision (ECCV)}, pages 532--547, 2018.

\bibitem{chaudhry2021using}
Arslan Chaudhry, Albert Gordo, Puneet Dokania, Philip Torr, and David Lopez-Paz.
\newblock Using hindsight to anchor past knowledge in continual learning.
\newblock In {\em Proceedings of the AAAI conference on artificial intelligence}, volume~35, pages 6993--7001, 2021.

\bibitem{chen2024saving}
Jinpeng Chen, Runmin Cong, Yuxuan Luo, Horace Ip, and Sam Kwong.
\newblock Saving 100x storage: Prototype replay for reconstructing training sample distribution in class-incremental semantic segmentation.
\newblock {\em Advances in Neural Information Processing Systems}, 36, 2024.

\bibitem{cheraghian2021semantic}
Ali Cheraghian, Shafin Rahman, Pengfei Fang, Soumava~Kumar Roy, Lars Petersson, and Mehrtash Harandi.
\newblock Semantic-aware knowledge distillation for few-shot class-incremental learning.
\newblock In {\em Proceedings of the IEEE/CVF conference on computer vision and pattern recognition}, pages 2534--2543, 2021.

\bibitem{cheraghian2021synthesized}
Ali Cheraghian, Shafin Rahman, Sameera Ramasinghe, Pengfei Fang, Christian Simon, Lars Petersson, and Mehrtash Harandi.
\newblock Synthesized feature based few-shot class-incremental learning on a mixture of subspaces.
\newblock In {\em Proceedings of the IEEE/CVF international conference on computer vision}, pages 8661--8670, 2021.

\bibitem{chi2022metafscil}
Zhixiang Chi, Li Gu, Huan Liu, Yang Wang, Yuanhao Yu, and Jin Tang.
\newblock Metafscil: A meta-learning approach for few-shot class incremental learning.
\newblock In {\em Proceedings of the IEEE/CVF conference on computer vision and pattern recognition}, pages 14166--14175, 2022.

\bibitem{de2021continual}
Matthias De~Lange, Rahaf Aljundi, Marc Masana, Sarah Parisot, Xu Jia, Ale{\v{s}} Leonardis, Gregory Slabaugh, and Tinne Tuytelaars.
\newblock A continual learning survey: Defying forgetting in classification tasks.
\newblock {\em IEEE transactions on pattern analysis and machine intelligence}, 44(7):3366--3385, 2021.

\bibitem{dong2021few}
Songlin Dong, Xiaopeng Hong, Xiaoyu Tao, Xinyuan Chang, Xing Wei, and Yihong Gong.
\newblock Few-shot class-incremental learning via relation knowledge distillation.
\newblock In {\em Proceedings of the AAAI Conference on Artificial Intelligence}, volume~35, pages 1255--1263, 2021.

\bibitem{dosovitskiy2020image}
Alexey Dosovitskiy, Lucas Beyer, Alexander Kolesnikov, Dirk Weissenborn, Xiaohua Zhai, Thomas Unterthiner, Mostafa Dehghani, Matthias Minderer, Georg Heigold, Sylvain Gelly, et~al.
\newblock An image is worth 16x16 words: Transformers for image recognition at scale.
\newblock {\em arXiv preprint arXiv:2010.11929}, 2020.

\bibitem{douillard2022dytox}
Arthur Douillard, Alexandre Ram{\'e}, Guillaume Couairon, and Matthieu Cord.
\newblock Dytox: Transformers for continual learning with dynamic token expansion.
\newblock In {\em Proceedings of the IEEE/CVF Conference on Computer Vision and Pattern Recognition}, pages 9285--9295, 2022.

\bibitem{goodfellow2014generative}
Ian Goodfellow, Jean Pouget-Abadie, Mehdi Mirza, Bing Xu, David Warde-Farley, Sherjil Ozair, Aaron Courville, and Yoshua Bengio.
\newblock Generative adversarial nets.
\newblock {\em Advances in neural information processing systems}, 27, 2014.

\bibitem{gu2023few}
Ziqi Gu, Chunyan Xu, Jian Yang, and Zhen Cui.
\newblock Few-shot continual infomax learning.
\newblock In {\em Proceedings of the IEEE/CVF International Conference on Computer Vision}, pages 19224--19233, 2023.

\bibitem{he2018exemplar}
Chen He, Ruiping Wang, Shiguang Shan, and Xilin Chen.
\newblock Exemplar-supported generative reproduction for class incremental learning.
\newblock In {\em BMVC}, page~98, 2018.

\bibitem{he2016deep}
Kaiming He, Xiangyu Zhang, Shaoqing Ren, and Jian Sun.
\newblock Deep residual learning for image recognition.
\newblock In {\em Proceedings of the IEEE conference on computer vision and pattern recognition}, pages 770--778, 2016.

\bibitem{hersche2022constrained}
Michael Hersche, Geethan Karunaratne, Giovanni Cherubini, Luca Benini, Abu Sebastian, and Abbas Rahimi.
\newblock Constrained few-shot class-incremental learning.
\newblock In {\em Proceedings of the IEEE/CVF Conference on Computer Vision and Pattern Recognition}, pages 9057--9067, 2022.

\bibitem{hu2018overcoming}
Wenpeng Hu, Zhou Lin, Bing Liu, Chongyang Tao, Zhengwei Tao, Jinwen Ma, Dongyan Zhao, and Rui Yan.
\newblock Overcoming catastrophic forgetting for continual learning via model adaptation.
\newblock In {\em International conference on learning representations}, 2018.

\bibitem{isele2018selective}
David Isele and Akansel Cosgun.
\newblock Selective experience replay for lifelong learning.
\newblock In {\em Proceedings of the AAAI Conference on Artificial Intelligence}, volume~32, 2018.

\bibitem{ji2023memorizing}
Zhong Ji, Zhishen Hou, Xiyao Liu, Yanwei Pang, and Xuelong Li.
\newblock Memorizing complementation network for few-shot class-incremental learning.
\newblock {\em IEEE Transactions on Image Processing}, 32:937--948, 2023.

\bibitem{jiang2021ib}
Jian Jiang, Edoardo Cetin, and Oya Celiktutan.
\newblock Ib-drr-incremental learning with information-back discrete representation replay.
\newblock In {\em Proceedings of the IEEE/CVF Conference on Computer Vision and Pattern Recognition}, pages 3533--3542, 2021.

\bibitem{kim2022warping}
Do-Yeon Kim, Dong-Jun Han, Jun Seo, and Jaekyun Moon.
\newblock Warping the space: Weight space rotation for class-incremental few-shot learning.
\newblock In {\em The Eleventh International Conference on Learning Representations}, 2022.

\bibitem{kingma2013auto}
Diederik~P Kingma and Max Welling.
\newblock Auto-encoding variational bayes.
\newblock {\em arXiv preprint arXiv:1312.6114}, 2013.

\bibitem{krizhevsky2009learning}
Alex Krizhevsky, Geoffrey Hinton, et~al.
\newblock Learning multiple layers of features from tiny images.
\newblock 2009.

\bibitem{kukleva2021generalized}
Anna Kukleva, Hilde Kuehne, and Bernt Schiele.
\newblock Generalized and incremental few-shot learning by explicit learning and calibration without forgetting.
\newblock In {\em Proceedings of the IEEE/CVF international conference on computer vision}, pages 9020--9029, 2021.

\bibitem{liu2022few}
Huan Liu, Li Gu, Zhixiang Chi, Yang Wang, Yuanhao Yu, Jun Chen, and Jin Tang.
\newblock Few-shot class-incremental learning via entropy-regularized data-free replay.
\newblock In {\em European Conference on Computer Vision}, pages 146--162. Springer, 2022.

\bibitem{mazumder2021few}
Pratik Mazumder, Pravendra Singh, and Piyush Rai.
\newblock Few-shot lifelong learning.
\newblock In {\em Proceedings of the AAAI Conference on Artificial Intelligence}, volume~35, pages 2337--2345, 2021.

\bibitem{radford2021learning}
Alec Radford, Jong~Wook Kim, Chris Hallacy, Aditya Ramesh, Gabriel Goh, Sandhini Agarwal, Girish Sastry, Amanda Askell, Pamela Mishkin, Jack Clark, et~al.
\newblock Learning transferable visual models from natural language supervision.
\newblock In {\em International conference on machine learning}, pages 8748--8763. PMLR, 2021.

\bibitem{rebuffi2017icarl}
Sylvestre-Alvise Rebuffi, Alexander Kolesnikov, Georg Sperl, and Christoph~H Lampert.
\newblock icarl: Incremental classifier and representation learning.
\newblock In {\em Proceedings of the IEEE conference on Computer Vision and Pattern Recognition}, pages 2001--2010, 2017.

\bibitem{rolnick2019experience}
David Rolnick, Arun Ahuja, Jonathan Schwarz, Timothy Lillicrap, and Gregory Wayne.
\newblock Experience replay for continual learning.
\newblock {\em Advances in Neural Information Processing Systems}, 32, 2019.

\bibitem{russakovsky2015imagenet}
Olga Russakovsky, Jia Deng, Hao Su, Jonathan Krause, Sanjeev Satheesh, Sean Ma, Zhiheng Huang, Andrej Karpathy, Aditya Khosla, Michael Bernstein, et~al.
\newblock Imagenet large scale visual recognition challenge.
\newblock {\em International journal of computer vision}, 115:211--252, 2015.

\bibitem{shi2021overcoming}
Guangyuan Shi, Jiaxin Chen, Wenlong Zhang, Li-Ming Zhan, and Xiao-Ming Wu.
\newblock Overcoming catastrophic forgetting in incremental few-shot learning by finding flat minima.
\newblock {\em Advances in neural information processing systems}, 34:6747--6761, 2021.

\bibitem{shin2017continual}
Hanul Shin, Jung~Kwon Lee, Jaehong Kim, and Jiwon Kim.
\newblock Continual learning with deep generative replay.
\newblock {\em Advances in neural information processing systems}, 30, 2017.

\bibitem{smith2023coda}
James~Seale Smith, Leonid Karlinsky, Vyshnavi Gutta, Paola Cascante-Bonilla, Donghyun Kim, Assaf Arbelle, Rameswar Panda, Rogerio Feris, and Zsolt Kira.
\newblock Coda-prompt: Continual decomposed attention-based prompting for rehearsal-free continual learning.
\newblock In {\em Proceedings of the IEEE/CVF Conference on Computer Vision and Pattern Recognition}, pages 11909--11919, 2023.

\bibitem{song2023learning}
Zeyin Song, Yifan Zhao, Yujun Shi, Peixi Peng, Li Yuan, and Yonghong Tian.
\newblock Learning with fantasy: Semantic-aware virtual contrastive constraint for few-shot class-incremental learning.
\newblock In {\em Proceedings of the IEEE/CVF Conference on Computer Vision and Pattern Recognition}, pages 24183--24192, 2023.

\bibitem{tao2020few}
Xiaoyu Tao, Xiaopeng Hong, Xinyuan Chang, Songlin Dong, Xing Wei, and Yihong Gong.
\newblock Few-shot class-incremental learning.
\newblock In {\em Proceedings of the IEEE/CVF Conference on Computer Vision and Pattern Recognition}, pages 12183--12192, 2020.

\bibitem{thengane2022clip}
Vishal Thengane, Salman Khan, Munawar Hayat, and Fahad Khan.
\newblock Clip model is an efficient continual learner.
\newblock {\em arXiv preprint arXiv:2210.03114}, 2022.

\bibitem{tian2023survey}
Songsong Tian, Lusi Li, Weijun Li, Hang Ran, Xin Ning, and Prayag Tiwari.
\newblock A survey on few-shot class-incremental learning.
\newblock {\em arXiv preprint arXiv:2304.08130}, 2023.

\bibitem{wah2011caltech}
Catherine Wah, Steve Branson, Peter Welinder, Pietro Perona, and Serge Belongie.
\newblock The caltech-ucsd birds-200-2011 dataset.
\newblock 2011.

\bibitem{wang2023comprehensive}
Liyuan Wang, Xingxing Zhang, Hang Su, and Jun Zhu.
\newblock A comprehensive survey of continual learning: Theory, method and application.
\newblock {\em arXiv preprint arXiv:2302.00487}, 2023.

\bibitem{wang2023attriclip}
Runqi Wang, Xiaoyue Duan, Guoliang Kang, Jianzhuang Liu, Shaohui Lin, Songcen Xu, Jinhu L{\"u}, and Baochang Zhang.
\newblock Attriclip: A non-incremental learner for incremental knowledge learning.
\newblock In {\em Proceedings of the IEEE/CVF Conference on Computer Vision and Pattern Recognition}, pages 3654--3663, 2023.

\bibitem{wang2022s}
Yabin Wang, Zhiwu Huang, and Xiaopeng Hong.
\newblock S-prompts learning with pre-trained transformers: An occam’s razor for domain incremental learning.
\newblock {\em Advances in Neural Information Processing Systems}, 35:5682--5695, 2022.

\bibitem{wang2023improving}
Zhengbo Wang, Jian Liang, Ran He, Nan Xu, Zilei Wang, and Tieniu Tan.
\newblock Improving zero-shot generalization for clip with synthesized prompts.
\newblock In {\em Proceedings of the IEEE/CVF International Conference on Computer Vision}, pages 3032--3042, 2023.

\bibitem{wang2022dualprompt}
Zifeng Wang, Zizhao Zhang, Sayna Ebrahimi, Ruoxi Sun, Han Zhang, Chen-Yu Lee, Xiaoqi Ren, Guolong Su, Vincent Perot, Jennifer Dy, et~al.
\newblock Dualprompt: Complementary prompting for rehearsal-free continual learning.
\newblock In {\em European Conference on Computer Vision}, pages 631--648. Springer, 2022.

\bibitem{wang2022learning}
Zifeng Wang, Zizhao Zhang, Chen-Yu Lee, Han Zhang, Ruoxi Sun, Xiaoqi Ren, Guolong Su, Vincent Perot, Jennifer Dy, and Tomas Pfister.
\newblock Learning to prompt for continual learning.
\newblock In {\em Proceedings of the IEEE/CVF Conference on Computer Vision and Pattern Recognition}, pages 139--149, 2022.

\bibitem{xiao2010sun}
Jianxiong Xiao, James Hays, Krista~A Ehinger, Aude Oliva, and Antonio Torralba.
\newblock Sun database: Large-scale scene recognition from abbey to zoo.
\newblock In {\em 2010 IEEE computer society conference on computer vision and pattern recognition}, pages 3485--3492. IEEE, 2010.

\bibitem{yang2021learnable}
Boyu Yang, Mingbao Lin, Binghao Liu, Mengying Fu, Chang Liu, Rongrong Ji, and Qixiang Ye.
\newblock Learnable expansion-and-compression network for few-shot class-incremental learning.
\newblock {\em arXiv preprint arXiv:2104.02281}, 2021.

\bibitem{yang2022dynamic}
Boyu Yang, Mingbao Lin, Yunxiao Zhang, Binghao Liu, Xiaodan Liang, Rongrong Ji, and Qixiang Ye.
\newblock Dynamic support network for few-shot class incremental learning.
\newblock {\em IEEE Transactions on Pattern Analysis and Machine Intelligence}, 45(3):2945--2951, 2022.

\bibitem{yang2023continual}
Yang Yang, Zhiying Cui, Junjie Xu, Changhong Zhong, Wei-Shi Zheng, and Ruixuan Wang.
\newblock Continual learning with bayesian model based on a fixed pre-trained feature extractor.
\newblock {\em Visual Intelligence}, 1(1):5, 2023.

\bibitem{yang2022neural}
Yibo Yang, Haobo Yuan, Xiangtai Li, Zhouchen Lin, Philip Torr, and Dacheng Tao.
\newblock Neural collapse inspired feature-classifier alignment for few-shot class-incremental learning.
\newblock In {\em The Eleventh International Conference on Learning Representations}, 2022.

\bibitem{yang2023neural}
Yibo Yang, Haobo Yuan, Xiangtai Li, Zhouchen Lin, Philip Torr, and Dacheng Tao.
\newblock Neural collapse inspired feature-classifier alignment for few-shot class incremental learning.
\newblock {\em arXiv preprint arXiv:2302.03004}, 2023.

\bibitem{zhang2021few}
Chi Zhang, Nan Song, Guosheng Lin, Yun Zheng, Pan Pan, and Yinghui Xu.
\newblock Few-shot incremental learning with continually evolved classifiers.
\newblock In {\em Proceedings of the IEEE/CVF conference on computer vision and pattern recognition}, pages 12455--12464, 2021.

\bibitem{zhang2023slca}
Gengwei Zhang, Liyuan Wang, Guoliang Kang, Ling Chen, and Yunchao Wei.
\newblock Slca: Slow learner with classifier alignment for continual learning on a pre-trained model.
\newblock In {\em Proceedings of the IEEE/CVF International Conference on Computer Vision}, pages 19148--19158, 2023.

\bibitem{zhang2023controlvideo}
Yabo Zhang, Yuxiang Wei, Dongsheng Jiang, Xiaopeng Zhang, Wangmeng Zuo, and Qi Tian.
\newblock Controlvideo: Training-free controllable text-to-video generation.
\newblock {\em arXiv preprint arXiv:2305.13077}, 2023.

\bibitem{zhao2021mgsvf}
Hanbin Zhao, Yongjian Fu, Mintong Kang, Qi Tian, Fei Wu, and Xi Li.
\newblock Mgsvf: Multi-grained slow vs. fast framework for few-shot class-incremental learning.
\newblock {\em IEEE Transactions on Pattern Analysis and Machine Intelligence}, 2021.

\bibitem{zhao2023few}
Linglan Zhao, Jing Lu, Yunlu Xu, Zhanzhan Cheng, Dashan Guo, Yi Niu, and Xiangzhong Fang.
\newblock Few-shot class-incremental learning via class-aware bilateral distillation.
\newblock In {\em Proceedings of the IEEE/CVF Conference on Computer Vision and Pattern Recognition}, pages 11838--11847, 2023.

\bibitem{zhou2022forward}
Da-Wei Zhou, Fu-Yun Wang, Han-Jia Ye, Liang Ma, Shiliang Pu, and De-Chuan Zhan.
\newblock Forward compatible few-shot class-incremental learning.
\newblock In {\em Proceedings of the IEEE/CVF conference on computer vision and pattern recognition}, pages 9046--9056, 2022.

\bibitem{zhou2023deep}
Da-Wei Zhou, Qi-Wei Wang, Zhi-Hong Qi, Han-Jia Ye, De-Chuan Zhan, and Ziwei Liu.
\newblock Deep class-incremental learning: A survey.
\newblock {\em arXiv preprint arXiv:2302.03648}, 2023.

\bibitem{zhou2022few}
Da-Wei Zhou, Han-Jia Ye, Liang Ma, Di Xie, Shiliang Pu, and De-Chuan Zhan.
\newblock Few-shot class-incremental learning by sampling multi-phase tasks.
\newblock {\em IEEE Transactions on Pattern Analysis and Machine Intelligence}, 2022.

\bibitem{zhou2022learning}
Kaiyang Zhou, Jingkang Yang, Chen~Change Loy, and Ziwei Liu.
\newblock Learning to prompt for vision-language models.
\newblock {\em International Journal of Computer Vision}, 130(9):2337--2348, 2022.

\bibitem{zhu2021self}
Kai Zhu, Yang Cao, Wei Zhai, Jie Cheng, and Zheng-Jun Zha.
\newblock Self-promoted prototype refinement for few-shot class-incremental learning.
\newblock In {\em Proceedings of the IEEE/CVF conference on computer vision and pattern recognition}, pages 6801--6810, 2021.

\bibitem{zhuang2023gkeal}
Huiping Zhuang, Zhenyu Weng, Run He, Zhiping Lin, and Ziqian Zeng.
\newblock Gkeal: Gaussian kernel embedded analytic learning for few-shot class incremental task.
\newblock In {\em Proceedings of the IEEE/CVF Conference on Computer Vision and Pattern Recognition}, pages 7746--7755, 2023.

\bibitem{zou2022margin}
Yixiong Zou, Shanghang Zhang, Yuhua Li, and Ruixuan Li.
\newblock Margin-based few-shot class-incremental learning with class-level overfitting mitigation.
\newblock {\em Advances in neural information processing systems}, 35:27267--27279, 2022.

\end{thebibliography}

\begin{IEEEbiography}[{\includegraphics[width=1in,height=1.25in,clip,keepaspectratio]{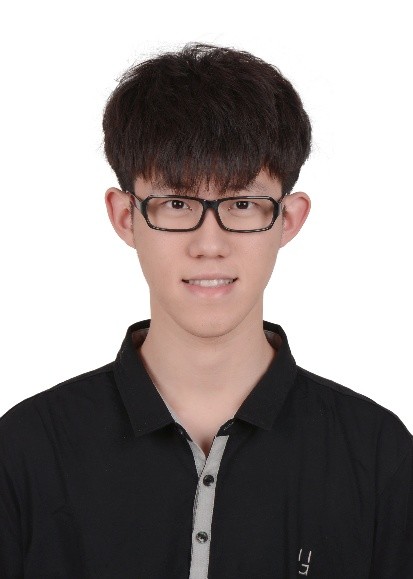}}]{Zitong Huang} is currently pursuing the Ph.D. degree with Harbin Institute of Technology, Harbin, China. His research interests include computer vision, deep learning, continual learning and object detection.
\end{IEEEbiography}

\begin{IEEEbiography}[{\includegraphics[width=1in,height=1.25in,clip,keepaspectratio]{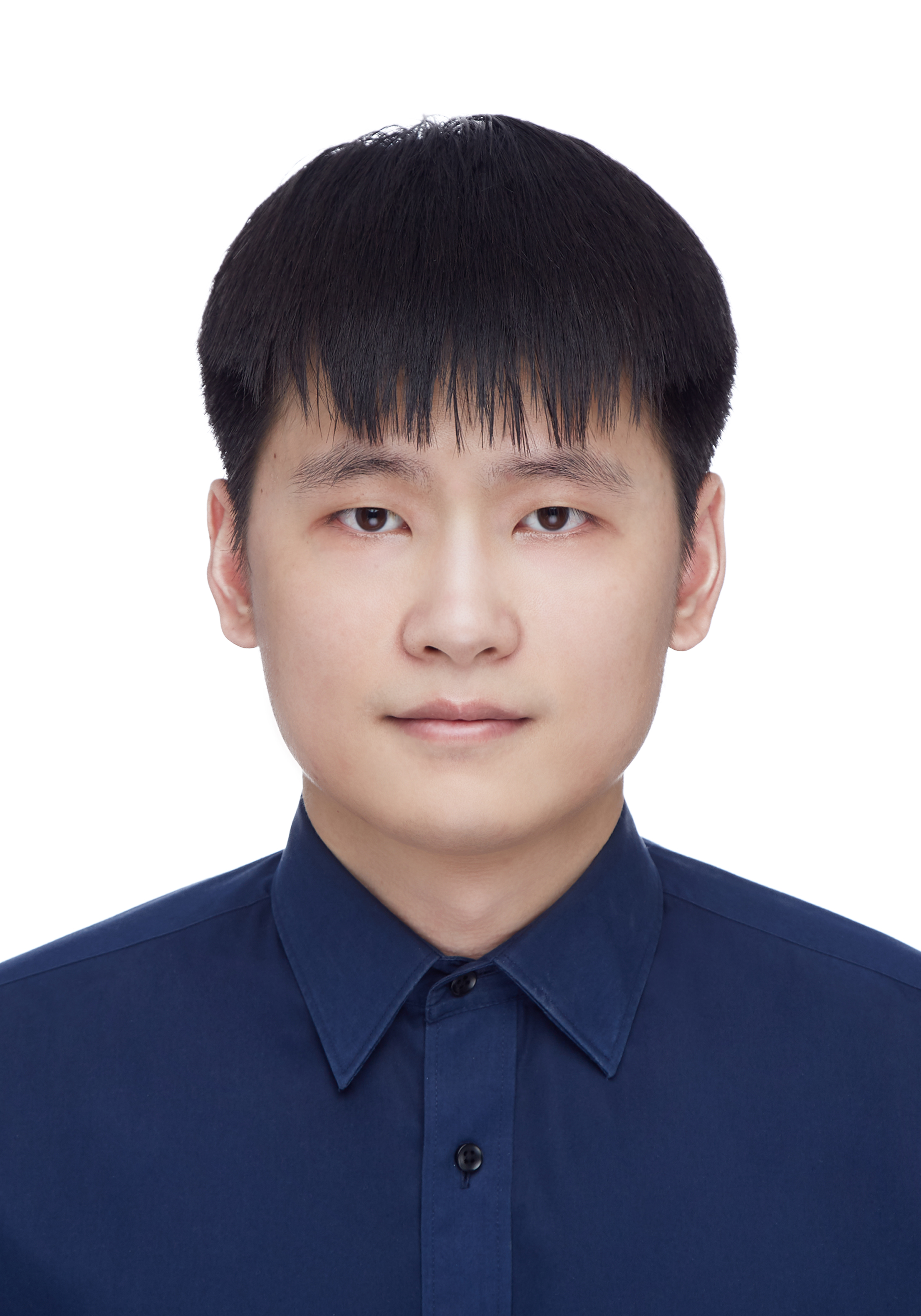}}]{Ze chen} is currently employed at Megvii Technology Limited, having graduated with a master's degree from Shanghai Jiao Tong University. His research interests encompass general object detection, the practical applications of deep learning models, and generative modeling.
\end{IEEEbiography}

\begin{IEEEbiography}[{\includegraphics[width=1in,height=1.25in,clip,keepaspectratio]{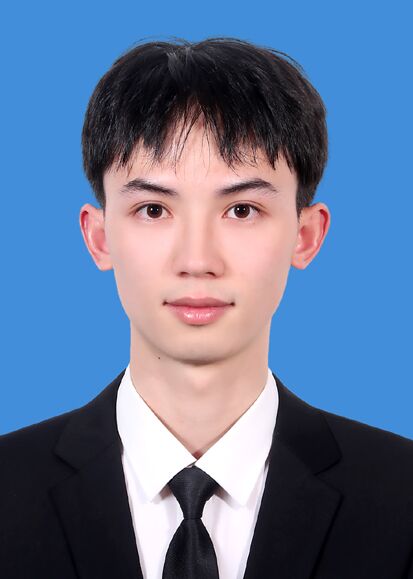}}]{Zhixing Chen} is currently pursuing a Bachelor’s degree in Robotics Engineering at Harbin Institute of Technology, China. His research interests include computer vision, artificial intelligence, and robotics.
\end{IEEEbiography}

\begin{IEEEbiography}[{\includegraphics[width=1in,height=1.25in,clip,keepaspectratio]{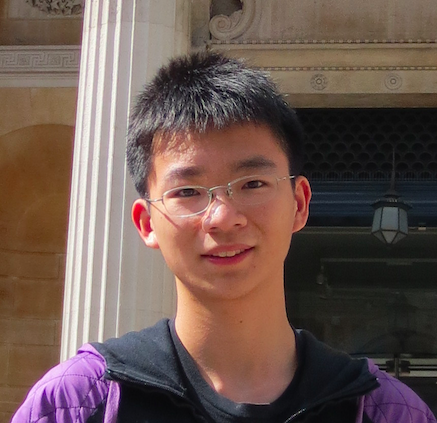}}]{Erjin Zhou} is the Director of the research group at Megvii Research Institute, responsible for building and leading the team to conduct research on facial and human body detection and recognition, portrait generation, general object detection, and action recognition technologies. Erjin Zhou also leads the team in the research and development of algorithm production tools. His research achievements have been applied to Megvii's cloud-based identity verification solution, as well as industry solutions for intelligent building access, smart phone AI solutions, and financial industry identity verification. 
\end{IEEEbiography}

\begin{IEEEbiography}[{\includegraphics[width=1in,height=1.25in,clip,keepaspectratio]{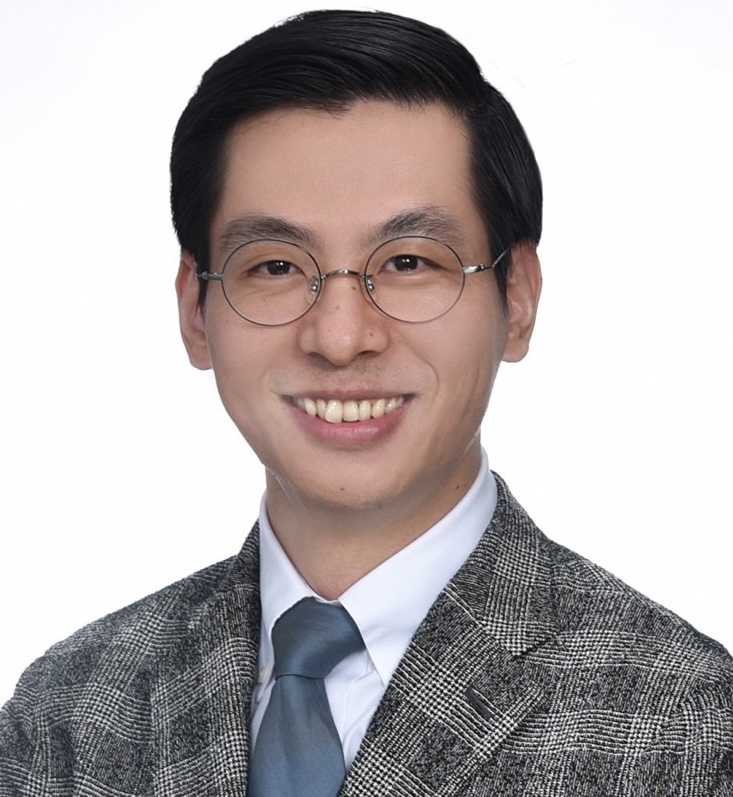}}]{Xinxing Xu} received his Ph.D. degrees from the Nanyang Technological University (NTU). He currently is a scientist with Institute of High Performance Computing (IHPC), Agency for Science, Technology and Research (A*STAR), Singapore. His research interests include computer vision, deep learning, and digital healthcare.
\end{IEEEbiography}

\begin{IEEEbiography}[{\includegraphics[width=1in,height=1.25in,clip,keepaspectratio]{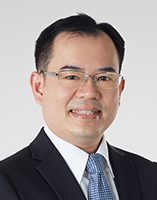}}]{Rick Siow Mong Goh} received his Ph.D. degree in electrical and computer engineering from the National University of Singapore in 2006. He is currently the Director of the Computing and Intelligence (CI) Department at A*STAR’s Institute of High Performance Computing (IHPC). He leads a team of over 80 scientists in performing world-leading scientific research, developing technology to commercialisation, and engaging and collaborating with industry. His research interests include artificial intelligence (AI), high performance computing, blockchain and federated learning.
\end{IEEEbiography}

\begin{IEEEbiography}[{\includegraphics[width=1in,height=1.25in,clip,keepaspectratio]{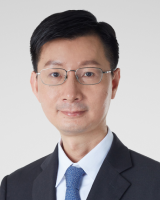}}]{Yong Liu} is the Deputy Department Director of Computing $\&$ Intelligence Department at Institute of High Performance Computing (IHPC), A*STAR, Singapore. He is also Adjunct Associate Professor at Duke-NUS Medical School, NUS and Adjunct Principal Investigator at Singapore Eye Research Institute (SERI). His research interests include computer vision and deep learning.
\end{IEEEbiography}

\begin{IEEEbiography}
[{\includegraphics[width=1in,height=1.25in,clip,keepaspectratio]{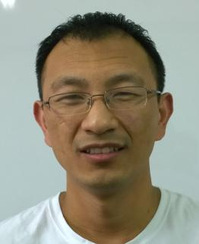}}]{Wangmeng Zuo} received the Ph.D. degree from the Harbin Institute of Technology in 2007. He is currently a Professor in the School of Computer Science and Technology, Harbin Institute of Technology. His research interests include image enhancement and restoration, image and face editing, object detection, visual tracking, and image classification. He has published over 100 papers in top tier journals and conferences. His publications have been cited more than 55,000 times. He also serves as Associate Editors for IEEE T-PAMI and IEEE T-IP.
\end{IEEEbiography}

\begin{IEEEbiography}
[{\includegraphics[width=1in,height=1.25in,clip,keepaspectratio]{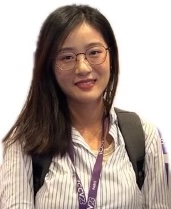}}]{Chun-Mei Feng}
	received the Ph.D. degree from Harbin Institute of Technology, Shenzhen. She is currently a research scientist in Institute of High Performance Computing (IHPC), Agency for Science, Technology and Research (A*STAR), Singapore. Her research interests include Federated learning, Medical image analysis, and Multi-modal foundation learning.
\end{IEEEbiography}

\vfill

\end{document}